\pdfoutput=1

\documentclass[11pt]{article}

\usepackage[preprint]{acl}

\usepackage{times}
\usepackage{latexsym}

\usepackage[T1]{fontenc}

\usepackage[utf8]{inputenc}

\usepackage{microtype}

\usepackage{inconsolata}

\usepackage{graphicx}
\usepackage{multirow}
\usepackage[normalem]{ulem}
\useunder{\uline}{\ul}{}
\usepackage{float}
\usepackage{booktabs}
\usepackage{array}
\usepackage{longtable}
\renewcommand{\thefootnote}{\fnsymbol{footnote}}

\usepackage{titlesec}
\titlespacing*{\paragraph}{0pt}{0.5ex plus 0.5ex minus .1ex}{0.5em}

\title{PIORS: Personalized Intelligent Outpatient Reception based on Large Language Model with Multi-Agents Medical Scenario Simulation}

\author{
    \textbf{Zhijie Bao\textsuperscript{\rm 1, \rm 2}\footnotemark[2]},
    \textbf{Qingyun Liu\textsuperscript{\rm 1}}\footnotemark[2],
    \textbf{Ying Guo\textsuperscript{\rm 3, \rm 4}},
    \textbf{Zhengqiang Ye\textsuperscript{\rm 5}},
    \textbf{Jun Shen\textsuperscript{\rm 3, \rm 6}},
    \textbf{Shirong Xie\textsuperscript{\rm 5}},
    \\
    \textbf{Jiajie Peng\textsuperscript{\rm 7}\footnotemark[1]},
    \textbf{Xuanjing Huang\textsuperscript{\rm 8}},
    \textbf{Zhongyu Wei\textsuperscript{\rm 1, \rm 2, \rm 9}\footnotemark[1]}
    \\
    \normalsize{\textsuperscript{1}School of Data Science, Fudan University, China,}
    \\
    \normalsize{\textsuperscript{2}MOE Laboratory for National Development and Intelligent Governance, Fudan University, China,}
    \\
    \normalsize{\textsuperscript{3}Shanghai Baoshan District Wusong Central Hospital, China,
    \textsuperscript{4}Zhongshan Hospital, Fudan University, China,}
    \\
    \normalsize{\textsuperscript{5}Eye \& ENT Hospital of Fudan University, China,}
    \normalsize{\textsuperscript{6}Zhongshan Hospital Wusong Branch, Fudan University, China,}
    \\
    \normalsize{\textsuperscript{7}School of Computer Science, Northwestern Polytechnical University, China,}
    \\
    \normalsize{\textsuperscript{8}School of Computer Science, Fudan University, China,}
    \\
    \normalsize{\textsuperscript{9}Research Institute of Automatic and Complex Systems, Fudan University, China}
    \\
    \normalsize{\texttt{\{zjbao24,qyliu21\}@m.fudan.edu.cn, jiajiepeng@nwpu.edu.cn}}\\
    \normalsize{\texttt{\{xjhuang,zywei\}@fudan.edu.cn}}
    \\
}

\begin{document}
\maketitle
\begin{abstract}

In China, receptionist nurses face overwhelming workloads in outpatient settings, limiting their time and attention for each patient and ultimately reducing service quality. 
In this paper, we present the \textbf{Personalized Intelligent Outpatient Reception System (PIORS)}. This system integrates an LLM-based reception nurse and a collaboration between LLM and hospital information system (HIS) into real outpatient reception setting, aiming to deliver personalized, high-quality, and efficient reception services. Additionally, to enhance the performance of LLMs in real-world healthcare scenarios, we propose a medical conversational data generation framework named \textbf{Service Flow aware Medical Scenario Simulation (SFMSS)}, aiming to adapt the LLM to the real-world environments and PIORS settings. We evaluate the effectiveness of PIORS and SFMSS through automatic and human assessments involving 15 users and 15 clinical experts. The results demonstrate that PIORS-Nurse outperforms all baselines, including the current state-of-the-art model GPT-4o, and aligns with human preferences and clinical needs. 
Further details and demo can be found at \emph{\url{https://github.com/FudanDISC/PIORS}}
\end{abstract}

\footnotetext[2]{The two authors contribute equal to this work.}
\footnotetext[1]{Corresponding authors.}

\section{Introduction}

\begin{figure*}[t]
    \centering
    \includegraphics[width=1\linewidth]{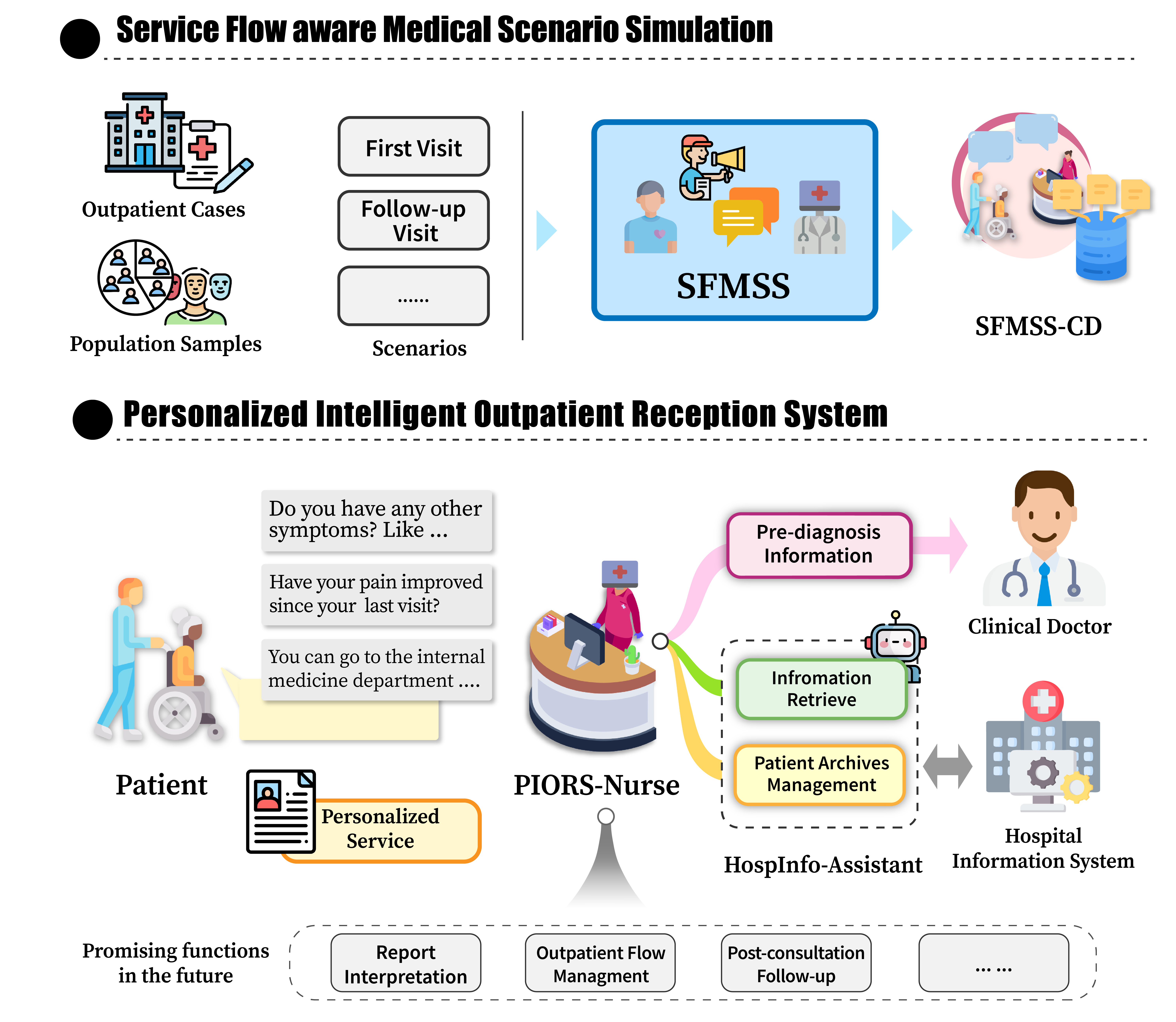}
    \caption{The overall framework of Personalized Intelligent Outpatient Reception System (PIORS). The upper section shows the framework of SFMSS, which supports the development of PIORS-Nurse. Bottom part is the details of PIORS, which involves four participants: a patient, a reception nurse (referred to as PIORS-Nurse), the clinical doctor and information assistant (referred to as HospInfo-Assistant). PIORS is integrated with the Hospital Information System (HIS), and the system's pipeline is depicted in this diagram.}
    \label{fig:method}
\end{figure*}

Outpatient reception play a crucial role in daily hospital workflow. Receptionist nurses are responsible for guiding patients to the appropriate departments and addressing their questions and concerns regarding general administration and primary care~\cite{Litchfield2017_rolereceptionist}.  An effective and empathetic supportive conversation will ease patients' anxiety and discomfort, improve their medical adherence, facilitate subsequent diagnosis and is the cornerstone of the overall positive experience~\cite{Kwame2021_nurse, Sharkiya2023_nursequality}. 

In 2023, China recorded an astounding 9.3 billion outpatient visits, representing a 13\% increase from the previous year~\cite{nhc2023_statistical}. This surge highlights the immerse burden on the healthcare system and ongoing challenges in the efficiency of hospital administration. Previous study indicate that receptionist nurses are required to address nearly one case per minute, communicating an average of 2,149.2 words per hour~\cite{wan_outpatient_2024}. This overwhelming workload restricts the time and attention available for each patient, leading to a limited quality of services provided. Moreover, patients often spend meaningless time in the waiting during the outpatient process, which could be utilized more productively to improve overall efficiency.


Given the remarkable capabilities and cost-effectiveness of large language models (LLMs), integrating them into the outpatient reception workflow presents a promising solution for these issues~\cite{Li2024_pre-consult, wan_outpatient_2024}. 

In this paper, we develop \textbf{Personalized Intelligent Outpatient Reception System (PIORS)}, which integrates an LLM-based reception nurse into real outpatient reception setting to deliver personalized, high-quality, and efficient reception services. The system is detailed in Figure 1, which include four participants, a patient, a reception nurse, the clinical doctor and a information assistant, and is integrated with the hospital information system (HIS).
The LLM nurse can access and maintain patients' information with assistance from a HospInfo-assistant, enabling responses that are relevant to patients' conditions and concerns. During the reception conversation, the LLM nurse gathers patients' medical information for improved department guiding. Following the reception process, details such as symptoms and medical history are updated in the patient archives and made available to clinical doctors, enhancing the efficiency of outpatient processes and reducing the workload of physicians. Furthermore, the LLM nurse is promising to integrate additional functions, such as report interpretation and examination planning, to offer more comprehensive and enriched outpatient reception services.

Additionally, to adapt the LLM to the real-world environments and PIORS settings, we propose a simulated-based framework named \textbf{Service Flow aware Medical Scenario Simulation (SFMSS)} for medical conversational data generation. This approach addresses the disconnection between traditional knowledge-centric LLM training methods and the complex, dynamic contexts encountered in practical settings~\cite{singhal_towards_2023, kung_performance_2023, liu_survey_2024, mehandru_evaluating_2024}. By using real-world data, simulating diverse patient interactions and incorporating service flow control, SFMSS supports the development of PIORS in a manner that more closely aligns with real-world scenarios.

Based on 2,400 Chinese hospital outpatient records, we develop the PIORS-Nurse via SFMSS. 
We conduct both automatic and human evaluations, involving 15 users and 15 clinical experts, to assess the effectiveness of our approach in outpatient reception. Results from the automatic evaluation demonstrate that our method outperforms all baselines, including the current state-of-the-art model GPT-4o, in terms of accuracy in department guiding and information-gathering capability. In the user evaluation, our method achieves a win or tie ratio of over 81\% compared to the best baseline, suggesting a better experience in real scenarios. Expert evaluation reveals that the PIORS-Nurse performs significantly better in inquiry capabilities and concise responses. 
Overall, By streamlining outpatient reception processes and improving communication with PIORS, our study contribute to a more efficient and patient-centered healthcare experience. Futher details and demo can be found at \emph{\url{https://github.com/FudanDISC/PIORS}}

\section{Personalized Intelligent Outpatient Reception System (PIORS)}

The service processes of PIORS involves four paticipants, a \textit{patient}, a\textit{nurse}, a \textit{information assistant} and the \textit{clinical doctor}, and is integrated with the \textit{hospital information system (HIS)}. Among these, nurse and information assistant are two LLM agents, specific to PIORS-Nurse (the development details is provided in \textsection\ref{sec:PIORS-Nurse-develop}) and HospInfo-Assistant respectively.

During the reception, PIORS-Nurse interacts with the patient to gather essential medical information and address their questions and concerns regarding administration and primary care. The medical information is used to guide them to the appropriate department or enhance the consultation of primary care. This interaction is enhanced by the HospInfo-Assistant, which make hospital information and patient archives available to PIORS-Nurse, enabling the provision of personalized and high-quality service. Following the reception process, details such as symptoms and medical history are updated in the patient archives and made available to clinical doctors, enhancing the efficiency of outpatient processes and reducing the workload of physicians.

\subsection{Workflow}
\paragraph{Patient Intake} The outpatient visit begins when a patient enters the hospital, where PIORS-Nurse receives the patient, conducts a pre-consultation to gather information about the chief complient and medical history, and provides appropriate department recommendation. Notably, for follow-up visits, PIORS-Nurse will raise retrieval requests for the corresponding initial consultation record and provides personalized inquiries based on this information. HospInfo-Assistant executes the retrieval query, obtains results from HIS, and forwards them to PIORS-Nurse.

\paragraph{Record Documentation and Handoff} At the conclusion of the conversation, PIORS-Nurse will summarize the dialogue to extract pre-diagnosis information, hand this information to subsequent clinical doctors, and send a record creation request to the HospInfo-Assistant. The assistant will then create a medical record in HIS, documenting the patient's basic information (name, gender, ID, and age), chief complaint, department assignment, and relevant medical history, including present illness history, past medical history, and drug allergy history.

\subsection{Duties of Nurse in PIORS}
%
\label{Nurse:role}
Reception nurses manage the patient intake process, responsible for guiding patients, addressing their basic inquiries and documenting their information. Based on the demand in outpatient reception scenario, we summarize the nurses' core responsibilities below.

\paragraph{Offer department guidance.} 
Ask patients for reasons coming to hospital and provide appropriate department recommendation based on their complaints. When the patient's description is too vague and precise department guidance is challenging, additional inquiry should be conducted.

\paragraph{Gather pre-diagnosis information.}
Further collect the detailed symptoms and medical history after the department recommendation is made. Mainly focus on chief complaint, history of present illness and past medical history. For follow-up visits, nurses should inquire about changes in symptoms, adherence to prescribed treatment, and other relevant aspects of the patient’s progress.

\paragraph{Administrative and primary care consultation.} 
Respond to patient queries about services, procedures and policies of hospital, provide clear and concise answers. Explain medical-related questions that patients may not understand, and address basic medical inquiries without offering specific medical advice.

\paragraph{Personalized service and Empathetic support.}
Greet patients and make them feel welcomed and comfortable. Ease patients when they feel anxiety or panic. Offer personalized service based on the patient's conditions and concerns.

\subsection{Composition of Nurse Agent}
\label{sec:nurse-framework}
We design a nurse agent composed of three key components: an interaction module for communication with patients, a query generation module for information retrieval, and a summarization module that extracts pre-diagnosis information from the dialogue.

\paragraph{Interaction Module}
The core component of the nurse agent, interacting directly with patients, address their questions and concerns regarding administration and primary care and guiding them to appropriate departments. It generates response based on the dialogue history, and will refer to retrieved information if given.



\paragraph{Query Generation Module}
This module first determines whether external information, such as hospital administrative information or patient archives, from HIS is needed and then generates a relevant query if required.

\paragraph{Summarization Module}
Iteratively extract symptoms and medical history from each turn of the dialogue, beginning with the first turn. For each subsequent turn, incorporate previously extracted information along with the new dialogue round to refine and expand the extracted information. Then summarize these information as medical record creation instruction and a pre-diagnosis report for subsequent clinical doctors.

\subsection{HospInfo-Assistant}
HospInfo-Assistant receives natural language instructions, extracts structured parameters, and calls APIs to interact with HIS. We prompt GPT-4-turbo to finish the process.

\paragraph{Patient Archives Management} 
The assistant manages patient archives through creating, selecting, updating, and deleting medical records. The assistant can identify the required operation and parameters from natural language input, then invoke the appropriate API to execute the operation. Personal information can be accessed by querying the medical records associated with the patient ID and is well-maintained.

\paragraph{Administrative Information Retrieval} 
When a administrative query is received, such as questions about the department locations or real-time physician schedules, HospInfo-Assistant will extracts key words and perform information retrieval in HIS.

\subsection{Promising Extensions in PIORS}
The system has the potential to expand to cover services during outpatient visits and post-visit care. The nurse agent in PIORS can be readily extended with additional functional modules, promising broad applicability across various tasks, including report explanation, outpatient flow management, and post-consultation follow-up. Further extend the accessibility of HospInfo-Assistant, automated payment, examination planning and booking will become possible.

\begin{figure*}[htbp]
    \centering
    \includegraphics[width=1\linewidth]{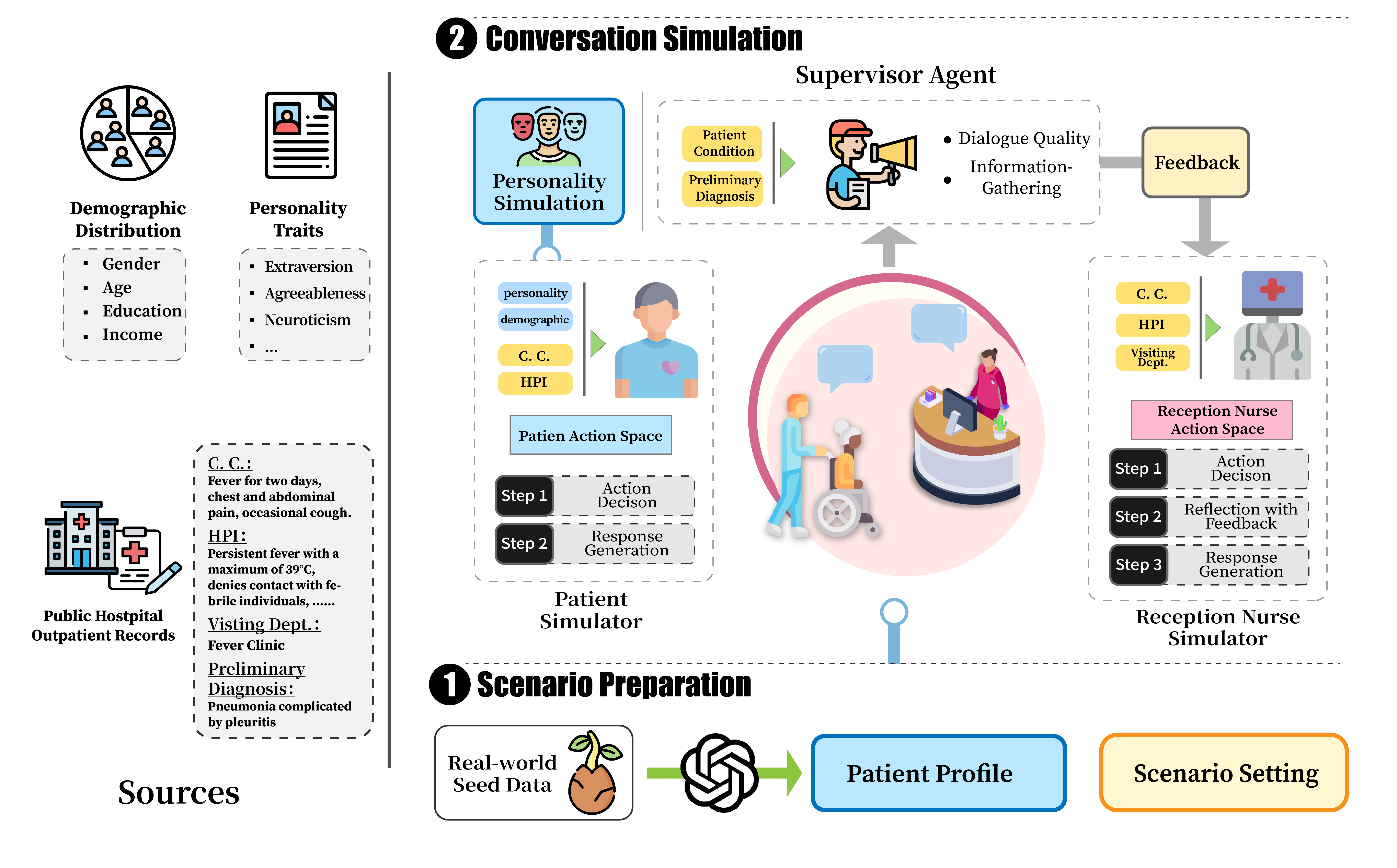}
    \caption{The details of Service Flow aware Medical Scenario Simulation (SFMSS). Left shows the data source guided the simulation, where the seed data is sampled from. Right part is the workflow of SFMSS, consisting of three main component: a patient simulator, a reception nurse simulator and a supervisor agent, with their internal response pipeline illustrated.}
    \label{fig:sfmss}
\end{figure*}
\renewcommand{\thefootnote}{\arabic{footnote}}
\section{Service Flow aware Medical Scenario Simulation (SFMSS)}
\label{sec:SFMSS}

Based on the needs of real-world medical scenarios, we propose a simulation-based data generation framework, named Service Flow aware Medical Scenario Simulation (SFMSS), to construct a high-quality dataset that captures the expected behavior of the reception nurse (\textsection\ref{Nurse:role}) and realistic patient interactions.

The SFMSS targets on the medical senario of outpatient reception and comprises three main LLM agents: a reception nurse simulator (referred to as ‘nurse’) and a patient simulator (referred to as ‘patient’) for conducting outpatient reception conversation, as well as a supervisor agent to refine the behavior of nurse. Simulators are built on the real-world data seed as input and employ the the role-playing capabilities of LLMs for dialogue generation. Given the modeling of actions and service flow, SFMSS ensures the conversations progress in alignment with medical objectives like department guiding in our scenario, facilitating goal-oriented decision-making. Appendix \ref{appendix:SFMSS-prompts} contains all the prompts of the SFMSS.

\subsection{Overall workflow}

\paragraph{Scenario Preparing:}
SFMSS utilize LLMs to generate scenario settings and patient profiles for agent initialization based on outpatient records, personality traits, and demographic characteristics. Detailed medical conditions and visiting department are provided to agents as references for information gathering and decision-making.

\paragraph{Conversation Simulation:}
The conversation starts by a \emph{patient} expresses the initial demand for healthcare services. Then followed by a multiple-turn dialogue between the \emph{nurse} and the \emph{patient}. In each turn, simulators choose an action and generate responses based on the action description and the context. At the end of each turn, the supervisor agent provides a feedback to the nurse. The conversation terminates when the patient selects the action that signals an end. 

\subsection{Reception nurse simulator}

Under outpatient reception scenario, the nurse conducts pre-consultation with patients, guiding them to correct department. To control the overall service flow in our scenario, we design a three-step pipeline and pre-define an action space, enabling the nurse to be aware of and follow the service flow.

\paragraph{Action space definition} 
Based on the primary tasks of the reception nurse outlining in \textsection\ref{Nurse:role}, we carefully define the seven actions of the nurse simulator with guidance from clinical experts: (1) Symptom inquiry, (2) Medical history inquiry, (3) Department recommendation, (4) Priority assistance, (5) Medical question answering, (6) Administrative question answering, (7) Conclusion and confirmation. Further descriptions of the actions, depending on the type of visit, are provided in Appendix \ref{appendix:action_space}.

\paragraph{Pipeline} In each round conversation, given the dialogue history, the response from the nurse simulator is generated through the following three-step process. (1) Action Decision. An action selector identifies the most suitable action from the predefined action space and retrieves corresponding description. (2) Reflection with Feedback. The agent receives feedback from the supervisor and make modification to the chosen action if needed. 
(3) Response Generation. The simulator generates the final response for this turn based on the dialogue history, medical conditions from outpatient case, specific of selected action and suggestions provided by supervisor. 

\subsection{Patient simulator}
We incorporate personality simulation into the patient simulator, and generate diverse patients with distinct communication styles and complex behavioral logic. This diversity contributes to better alignment with real-world scenarios.
\paragraph{Personality simulation} We derive the Big Five personality traits and demographic characteristics from real-world to simulate characters of the patient. This include gender, age, incoming level, education level and openness to experience, conscientiousness, extraversion, agreeableness, neuroticism. Patient profile is generated through the following two-step process. (1) Sampling. We first sample all 9 attributes from real data. For personality, We provide a range of 3 to 8 adjectives for each Big Five trait (Appendix \ref{appendix:personality}) identified as high or low (high/low/moderate for each traits). We then select two adjectives for each trait not identified as moderate, to represent their characteristics. For demographic characteristics, we maintain the original form of data. (2) Patient Profile Generating. We prompt GPT-4o to generate a patient profile that covers the patient's information, personality traits and behavioral preferences with natural language. Profile will affect communication styles, action decision and emotional expression of the patient during interactions.






\paragraph{Action space definition} We pre-define five types of actions for a patient: (1) Expressing Needs, (2) Information Feedback, (3) Mention other topic, (4) Inquiry, (5) Ending the Conversation. Further descriptions of the actions are provided in Appendix \ref{appendix:action_space}.

\paragraph{Pipeline} The implementation of patient simulator starting with the profile generation. In each turn of a conversation, a response is generated following two steps.  (1) Action Decision. An action selector identifies the most suitable action from the predefined action space and retrieves the description of this selected action. Profile of patient will affect the decision of action. (2) Response Generation. The simulator generates the final response for this turn based on the dialogue history, medical conditions from outpatient case, patient profile and specific of selected action. 

\subsection{Supervisor agent}

The supervisor agent is designed to oversee the overall quality of dialogue and the completeness of information gathering. After each patient response, it provides two suggestions to the nurse agent. The supervisor agent comprises two sub-agents.

\paragraph{Dialogue quality.} The dialogue quality supervisor mainly focus on monitoring patient emotions and the effectiveness of the dialogue. When the latest input from the patient indicates clear dissatisfaction or when there are multiple rounds of repetitive/ineffective dialogue, appropriate suggestions are given to the nurse agent.

\paragraph{Information gathering.} The information gathering supervisor consists of three components: a memory bank, an information extractor and a suggestion generator. Whenever the nurse and a patient complete a round of dialogue, the information extractor extract new patient information given in the new round and add them to the memory bank. Then, the suggestion generator will compare the known memory bank and the true patient profile, and determine if the information collected is complete. If it is not completed, it will provide recommendations and corresponding action for further information collection.

\section{PIORS-Nurse Development}




\subsection{Source Dataset}\label{sec:data}
\paragraph{Real-world Outpatient Records.} We collect the year 2023-2024 outpatient record data from a public hospital, which consists of over 750k examples covering 174 departments. Each record primarily consists of the following fields: chief complaint, present illness history, past history, department and preliminary diagnosis. All field names and their detailed descriptions are in the Appendix \ref{appendix:record_details}. The real-world data is noisy and often of low quality, so we filter the records and retained a smaller high-quality dataset containing 25k examples from 36 departments. Then we extract 2,400 records as the \textbf{training set} and 500 records as the \textbf{test set} from the filter dataset through stratified sampling by department. The training set is used to construct the Supervised Fine-Tuning (SFT) dataset, while the test set is used to assess the performance of different models when serving as nurses.

\paragraph{Distribution of Real Patients} The demographic distribution of patients is from Analysis Report of National Health Services Survey in China, 2018 \cite{nhc2018}. For the BigFive personality traits, we utilize the dataset collected over 1M online questionnaire answers to 50 personality items\footnote{\url{https://www.kaggle.com/datasets/tunguz/big-five-personality-test}} to represent the real distribution. To transform the data from questionnaire answers to high, low or moderate levels of each trait, we sum up all scores (positive score for high description and negative score for low description) and classify the questionnaire based on the distance between the sum and the median of the whole dataset.

\subsection{SFMSS-CD}
\label{sec:SFMSS-CD}
We divide the training set into two parts: 2,000 records for first visit simulation and 400 records for follow-up visit simulation.

We utilize SFMSS, based on GPT-4o, to construct a simulated conversation dataset for outpatient reception, named \textbf{SFMSS-CD}. To accommodate different types of visits, we define two distinct action space descriptions for for the nurse and patient simulators, each designed to reflect their respective behaviors in first and follow-up visit scenarios. The detailed descriptions of each action are provided in Appendix\ref{appendix:action_space}. SFMSS-CD is composed of two parts: 2,000 first-visit conversations generated through SFMSS with the first visit action space definition, and 400 follow-up conversations generated using SFMSS with the corresponding follow-up action space definition.

Together, SFMSS-CD is a high-fidelity simulated conversation dataset covering 36 departments and various contexts, comprising a total of 2,400 samples.

\label{sec:PIORS-Nurse-develop}
\subsection{Models used in PIORS-Nurse}
PIORS-Nurse consists of three components, interaction module, query generation module and summarization module. For the last two modules, we leverage the capabilities of general-purpose LLMs: the query generation module utilizes GPT-4o-mini, while the summarization module employs GPT-4o. For interaction module, the core part, we develop a model specifically adapted for real-world outpatient reception scenarios, capable of fulfilling the specified responsibilities of reception nurse. 

\subsection{Details of Implementation}
\label{SFMSS-Nurse}
Interaction module is fine-tuned on top of Qwen2-7B-Instruct~\cite{qwen2}. Qwen2-7B-Instruct is an open-source LLM with 7 billion parameters, trained and further instruction-tuned on a large-scale dataset that encompasses up to 7 trillion tokens. It exhibits ideal performance, especially in Chinese.

The SFT dataset is constructed by integrating 1,000 general-domain conversation and instruction samples with SFMSS-CD, comprising 3,400 samples in total. The general-domian data consists of  consists of 500 samples from sharegpt\_gpt4\footnote{\url{https://huggingface.co/datasets/shibing624/sharegpt_gpt4}}, and 500 samples from CQIA-Subset\footnote{\url{https://huggingface.co/datasets/m-a-p/COIG-CQIA}}.


For training, we complete the full-parameter SFT stage on 8*A100 GPUs, with the hyperparameters setting as follows: global batch size of 32, learning rate of 1e-5 with cosine scheduler, 3 epochs, maximum sequence length of 4096, warm up steps of 20 and with 0.05 weight decay.

\section{Automatic Evaluation}
\label{auto}


To evaluate the core component of our system, PIORS-Nurse, we introduce an automatic dynamic evaluation pipeline to assess the performance of different models playing the role as reception nurse, using the test set as simulated patient profiles.

\subsection{Evaluation Pipeline}
The whole pipeline includes two parts: dialogue simulation and quality evaluation. First, prompt the selected model to play the role of nurse, then interact with the patient simulator defined in SFMSS under the outpatient reception scenario and derive the simulated dialogue. One conversation ends when patient simulator picks the ending action or the dialogue exceeds 10 rounds. Second, assess the model's performance by evaluating the quality of simulated dialogues. 


\subsection{Evaluation Metrics}
Based on the core responsibilities of nurses and the needs in outpatient reception scenarios, we design the following four dimensions to assess the performance of different models.

\textbf{Accuracy}: We evaluate the accuracy by comparing the department recommended by the model with the ground truth label given by human doctors.

\textbf{Efficiency}: The efficiency can be measured by length of the patient-nurse dialogue. This includes two metrics: \textbf{Average Turn Number} and \textbf{Average Turn Length}. 

\textbf{Information Gathering Ability}: Collecting symptoms and medical history in advance can facilitate subsequent doctor diagnoses and improve the accuracy of triage. From this perspective, we introduce \textbf{Info Score} for information gathering, and prompt GPT-4o as evaluator to provide a 5-point score, given the true patient profile for reference.

\textbf{Overall Performance}: To assess the overall performance of a nurse model, we prompted the GPT-4o as evaluator to provide a 5-point \textbf{Overall Score} focusing on whether the core responsibilities are fulfilled and the quality of task completion. 


\subsection{Baselines}
We first directly prompt current general LLMs GPT-4o, Qwen2-7B-Instruct and Meta-Llama-3-8b-instruct to play the role of reception nurse~\cite{qwen2,openaigpt4o,llama3modelcard}. We also prompt HuaTuoGPT2-13B~\cite{chen2023huatuogpt}, a model specialized in medical domain, to show the gap between traditional training knowledge and complex real-world settings.

\subsection{Ablation study}

For further study on the effectiveness of the workflow and functional design of the nurse in PIORS, we introduce an ablated model, service-flow-ablated nurse (SF-ablated nurse), which is fine-tuned with same training settings describe in \textsection\ref{SFMSS-Nurse}, utilizing the data generated solely based on the LLM's parametric knowledge. In this approach, GPT-4o is directly prompted as a nurse simulator given the true department, without a pre-defined action space or supervisor agent, while the patient simulator and scene description remains unchanged. Prompts are provided in Appendix \ref{appendix:prompt_baseline}.

\subsection{Overall Results}



\begin{table*}[htbp]
\centering
\begin{tabular}{@{}ccccccc@{}}
\toprule
Method                           & Model                                                                    & Accuracy       & \begin{tabular}[c]{@{}c@{}}Overall\\ Score\end{tabular} & \begin{tabular}[c]{@{}c@{}}Info\\ score\end{tabular} & \begin{tabular}[c]{@{}c@{}}Average\\ Turn Number\end{tabular} & \begin{tabular}[c]{@{}c@{}}Average \\ Turn Length\end{tabular} \\ \midrule
\multirow{4}{*}{Directly Prompt} & \textbf{GPT-4o}                                                          & 0.717          & 3.83                                                    & 2.16                                                 & 3.54                                                          & 207.98                                                  \\
                                 & \textbf{Qwen2-7B}                                                        & 0.634          & 3.65                                                    & 2.28                                                 & 4.22                                                          & 336.40                                                         \\
                                 & \textbf{Llama3-8B}                                                        & 0.401          & 3.24                                                    & {\ul 2.65}                                           & 4.44                                                          & 678.14                                                         \\
                                 & \textbf{HuatuoGPT2-13B}                                                  & 0.501          & 3.25                                                    & 2.17                                                 & 3.57                                                          & 258.38                                                         \\ \midrule
\multirow{2}{*}{Fine-tuned}      & \textbf{\begin{tabular}[c]{@{}c@{}}SF-ablated\\ nurse\end{tabular}} & {\ul 0.786}    & {\ul 3.92}                                              & 2.20                                                 & {\ul 3.37}                                                    & {\ul 202.55}                                                   \\
                                 & \textbf{PIORS-Nurse}                                                      & \textbf{0.822} & \textbf{4.01}                                           & \textbf{3.01}                                        & \textbf{3.22}                                                 & \textbf{139.54}                                                \\ \bottomrule
\end{tabular}
\caption{Overall results of automatic evaluation. The highest score/shortest length is highlighted in bold, while the second highest/shortest is underlined.}
\label{overall_result}
\end{table*}

The overall results are shown in Table \ref{overall_result}. PIORS-Nurse ranks the first in all performance metrics, and has the shortest average turn number and average turn length.

\textbf{PIORS contributes to more accurate department guidance.} 
PIORS-Nurse demonstrates a significant improvement in accuracy compared to baselines without fine-tuning, demonstrating an 18\% increase relative to the Qwen2-7b backbone. Moreover, PIORS-Nurse achieves higher accuracy compared to the SF-ablated nurse, emphasizing that the pre-designed functions and workflow in PIORS contribute to better department guidance. This also demonstrates that the heuristic knowledge is successfully embedded in LLMs through the pre-defined action space and the proposed supervisory agent.

\textbf{Enhanced Information Gathering in Patient Intake Process by PIORS} PIORS-Nurse outperforms all baseline models by a large margin in Info Score metric. This underscores the model's enhanced pre-diagnosis information-gathering abilities. Distribution of Info Score is in Appendix \ref{appendix:add_results}.

\textbf{Efficient outpatient reception services via PIORS.} As shown in Table\ref{overall_result}, dialogues guided by PIORS-Nurse tend to have the shortest turn length, with over 70 characters fewer compared to the SF-ablated nurse baseline, and the shortest rounds. The total conversation length is significantly shorter compared to all baselines. These results ensure the efficiency and seamlessness communication between PIORS-Nurse and patients. 


\section{Human Evaluation}
To ensure high-quality evaluations from those with significant medical experience (experts) and from the population who may engage with PIORS in practice (users), we conducted separate assessments for users and experts.
\subsection{User Study}
We recruit 15 volunteers to participate in the study, and randomly sample 20 records from test set as patient profiles. Each participant was assigned to simulate the 20 patients sequentially based on given profiles and engage in conversations with PIORS-Nurse, GPT-4o, and baseline, respectively. After completing the three conversations in each iteration, they made blind pairwise comparisons of PIORS-Nurse with other two model, selecting from the options: "A is better than B (Win)", "about the same (Tie)", or "B is better than A (Loss)".


\begin{figure}[htbp]
    \centering
    \includegraphics[width=1\linewidth]{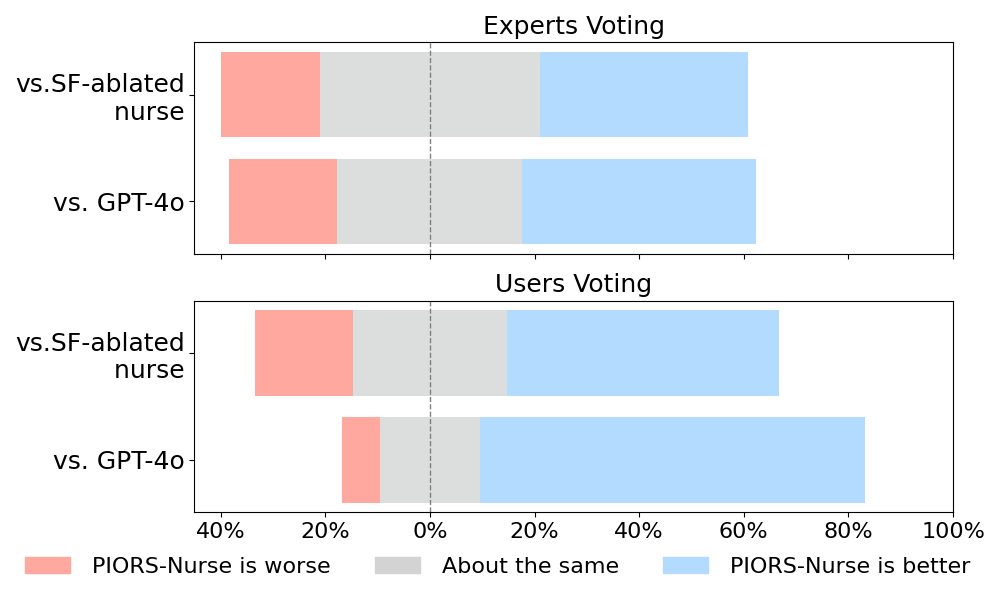}
    \caption{Performace of PIORS-Nurse compared to GPT-4o and SF-ablated nurse. X-axis: \% of examples voted by experts or users for a specific option, y-axis: the comparison model.}
    \label{fig:human_rate}
\end{figure}

Figure \ref{fig:human_rate} (lower) depicts the distribution of user comparisons. PIORS-Nurse has a win rate up to 73\% and a win-or-tie rate higher than 90\% compared to GPT-4o. Even compared to the fine-tuned SF-ablated nurse, our model still demonstrates a win-or-tie rate of over 81\%. This result indicates that PIORS-Nurse aligns more closely with human preferences and can better handle the complex real-world scenarios of outpatient reception.

\subsection{Expert Evaluation}
We recruit 15 experts, who are currently graduate or PhD students in clinical psychology, or have worked for more than 2 months in this field. 
Each expert was randomly assigned with 20 samples, each consists of three dialogues from the same simulated patient and different reception nurses role-played by PIORS-Nurse, GPT-4o, and SF-ablated nurse. Experts were asked to make two pairwise comparisons between PIORS-Nurse and other two models blindly, and evaluate the fidelity of simulated patients.


The comparison options for nurse are the same as those used in the user study. Fidelity is assessed using four levels: "Extremely High", "High", "Moderate" and "Low". The detailed definitions of patient fidelity are in Appendix \ref{fidelity_levels}.


Figure \ref{fig:human_rate} (upper) shows the distribution of expert comparisons. In 80\% samples, experts voted PIORS-Nurse has better or comparable performance compared to SF-ablated nurse. And the ratio is even higher when compared to GPT-4o. Furthermore, 12 of 15 clinical experts believed that PIORS-Nurse performed the best overall in the evaluation of the 20 samples.

The experts highly commended PIORS-Nurse for its proactive inquiry capabilities and concise responses. They noted that PIORS-Nurse's reasoning more closely aligns with that of real-world nurses. Furthermore, they observed that PIORS-Nurse remains focused on department triage throughout the conversation, without being easily diverted by patient input.

\section{Further Analysis}
In this section, we provide a more detailed analysis of the experiment results. Further demonstrate SFMSS's ability to simulate complex real-world scenarios.
\subsection{Fidelity of Patient Simulation}
\begin{figure}[htbp]
    \centering
    \includegraphics[width=1\linewidth]{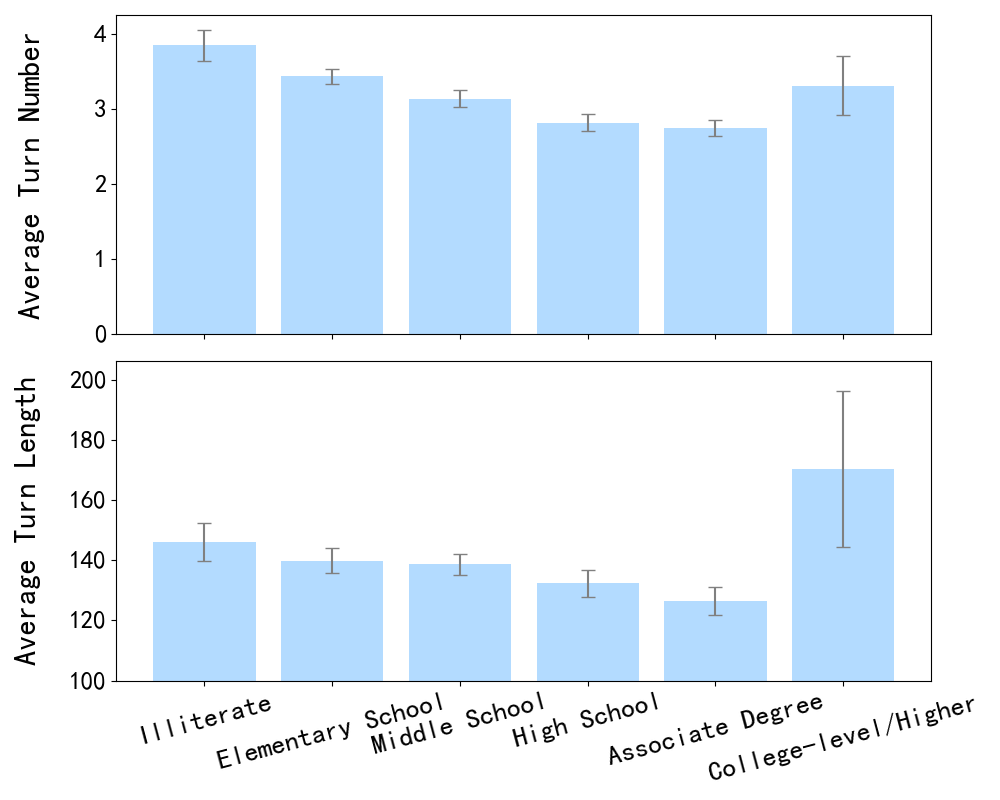}
    \caption{The average turn number and average turn length of simulated patient-nurse dialogues grouped by education level.}
    \label{fig:edu_stat}
\end{figure}

\textbf{Variation in Behavior} 
We use education level as an example to analyse the behavioral differences in simulated patients across various attribute settings. As shown in Figure \ref{fig:edu_stat}, from illiterate to associate degree, the conversations become shorter and more efficient. This aligns with the impact of education level: higher levels of education enable patients to communicate more effectively with triage nurses. However, when patients are assigned a college level or higher, the average number of characters per turn increases significantly, and the dialogues tend to involve more rounds. This may be attributed to two factors: 1) The small sample size for patients with the highest education level. 2) Patients with higher education levels are more likely to express their own opinions rather than simply following the guidance provided by reception nurses.

\textbf{Clinical experts noted that most simulated patients are indistinguishable from or close to real patients.} Notably, during dynamic interactions, the patient's behavior is influenced by the actions of the nurse(Table \ref{patient_rate}). Under blinded conditions, PIORS-Nurse achieves the highest level of fidelity, suggesting that our approach is most closely aligned with real-world scenarios.

\setlength{\tabcolsep}{4pt}
\begin{table}[htbp]
\centering
\begin{tabular}{@{}lcccc@{}}
\toprule
Nurse                                                          & \begin{tabular}[c]{@{}c@{}}Extremely\\ High\end{tabular} & High    & Moderate & Low    \\ \midrule
\textbf{GPT-4o}                                                & 29.3\%                                                  & 47.3\% & 20.0\%  & 3.3\% \\
\textbf{\begin{tabular}[c]{@{}l@{}}SF-ablated\\ nurse\end{tabular}}                                              & 31.0\%                                                  & 50.3\% & 14.3\%  & 4.3\% \\
\textbf{\begin{tabular}[c]{@{}l@{}}PIORS-\\ Nurse\end{tabular}} & 37.7\%                                                  & 47.0\% & 12.0\%  & 3.3\% \\
\textbf{Overall}                                               & 32.7\%                                                  & 48.2\% & 15.4\%  & 3.7\% \\ \bottomrule
\end{tabular}
\caption{Experts voting for fidelity of patient simulation with different models playing role of nurse. Extremely high means can't distinguish from human. High means closely resemble real patient behaviour.}
\label{patient_rate}
\end{table}

\subsection{Results of Different Patient Simulation}
\begin{figure}[htbp]
    \centering
    \includegraphics[width=1\linewidth]{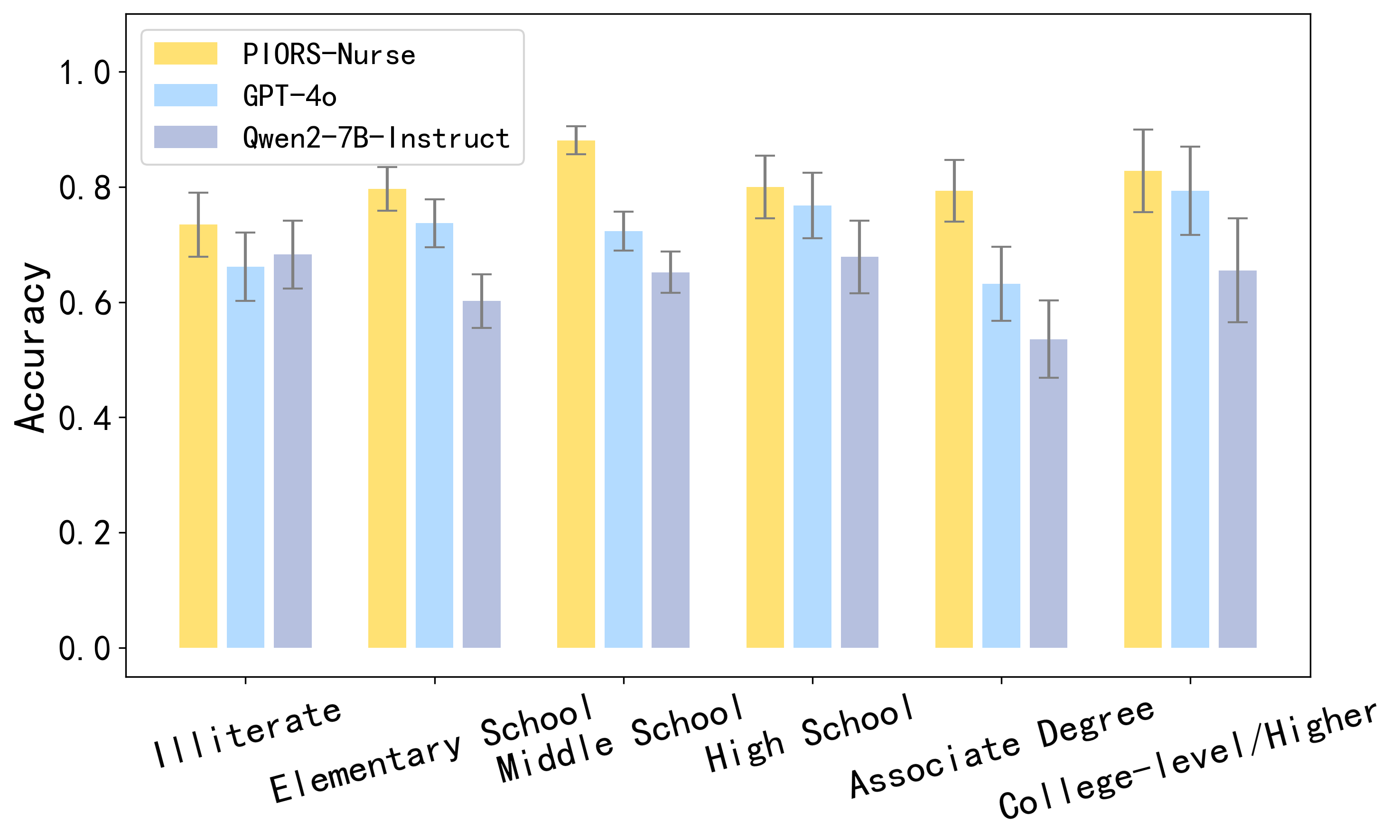}
    \caption{Accuracy grouped by education level. }
    \label{fig:edu_acc}
\end{figure}
\textbf{Education Level}
From the results of GPT-4o in Figure \ref{fig:edu_acc}, we can observe that, except for the associate degree, the accuracy of department guiding increases with higher education levels. This is consistent with common sense. 
The exception may stem from the misalignment of the term "Associate Degree" between GPT-4o's internal knowledge and its actual meaning in Chinese context. Figure \ref{fig:edu_acc} further illustrates the improvements from Qwen2-7B to PIORS-Nurse, especially in higher education level. After embedding the service flow and designed duties through SFT, the model can better addressing tasks in real-world scenarios, where primary and middle school education levels are most prevalent.

\textbf{BigFive Personality Traits}  
\begin{figure}
    \centering
    \includegraphics[width=1\linewidth]{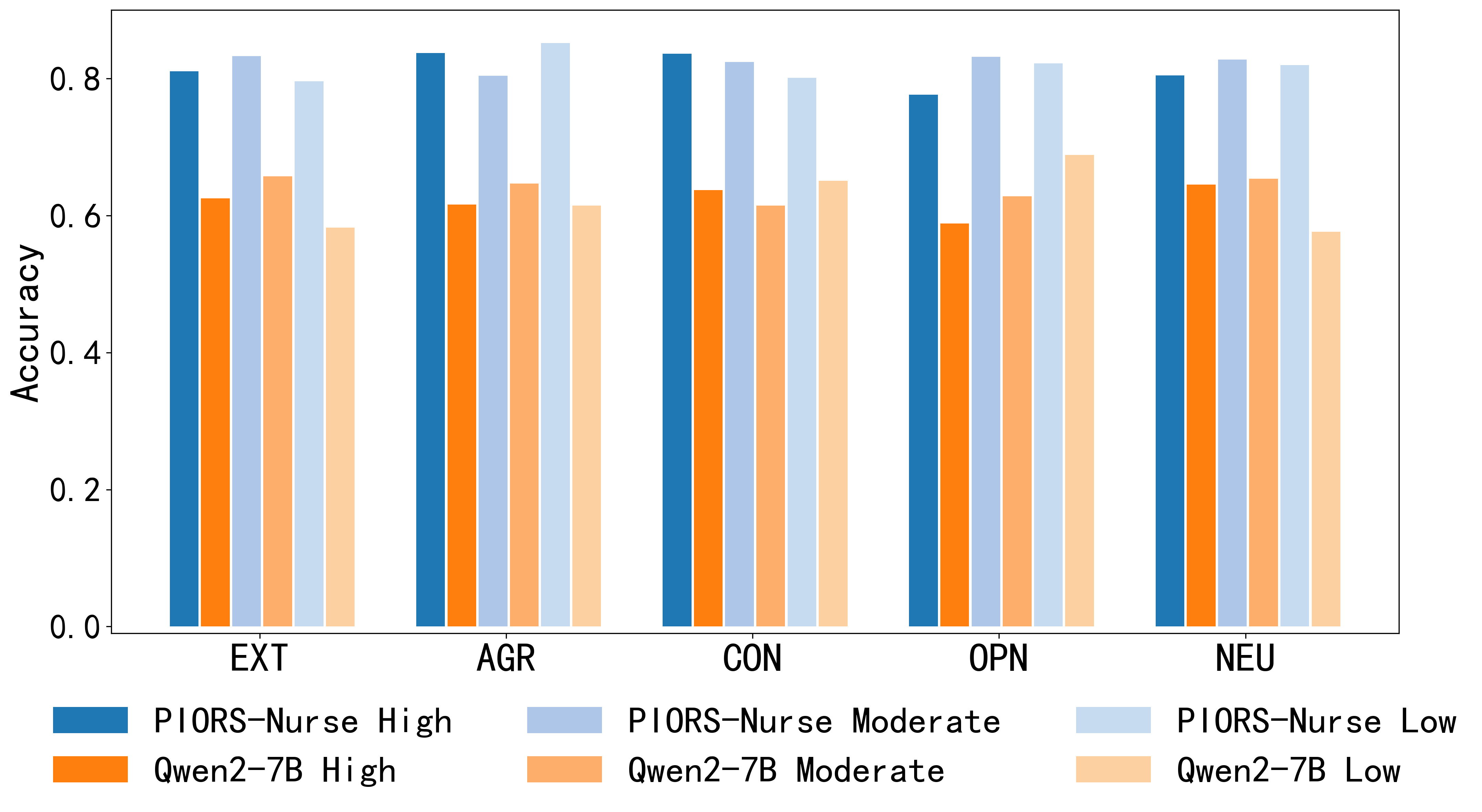}
    \caption{Accuracy grouped by BigFive personality traits. EXT refers to Extraversion, AGR refers to Agreeableness, CON refers to Conscientiousness, OPN refers to Openness to Experience and NEU refers to Neuroticism.}
    \label{fig:big5_acc}
\end{figure}
For Qwen2-7B, patients with specific personalities, such as high extraversion, low openness to experience and high neuroticism are more easy to handle (Figure \ref{fig:big5_acc}). Such preference can be explained: These personalities tend to provide more information (high extraversion and high neuroticism) or introduce fewer off-topic discussions(low openness to experience). 
After training, PIORS-Nurse exhibits a more balanced performance across various personality traits, allowing it to better adapt to complex healthcare scenarios.




\section{Related Works}
\paragraph{Medical LLMs and its Application}
Natural Language Processing (NLP) has found extensive application in the medical field~\cite{DXformer_Chen, bench_Chen, Zhong2022, wei-etal-2018-task}. With the advent of large language models (LLMs), significant advancements have been made in developing models specially for healthcare-domain. Models like Med-PaLM2~\cite{singhal_towards_2023}, DISC-MedLLM~\cite{bao_disc-medllm_2023}, HuatuoGPT-2~\cite{chen2023huatuogpt}, Med-Gemini~\cite{saab_capabilities_2024}, Aqulia-Med~\cite{zhao_aqulia-med_2024} are developed using different datasets, techniques, and frameworks. These efforts primarily focus on enhancing the medical knowledge embedded within language models and often fall short of effectively addressing the complexities of real-world scenarios. 
Recent works have attempted to apply LLMs to a variety of real-world clinical scenarios, such as radiology and ophthalmology~\cite{liu_evaluating_2023, gao_ophglm_2023, waisberg_gpt-4_2023, Yang2022}. The works of Wan et al.~\cite{wan_outpatient_2024} and Liu et al.~\cite{eval_triage_liu} has demonstrated the promising prospects of LLMs in outpatient reception scenarios. Despite these advancements, the challenges posed by the intricacies of real patient interactions remain insufficiently addressed in current research.

\paragraph{Scenario Simulation in Healthcare}

Recent developments indicate that agents powered by LLMs can resolve complex tasks through human-like actions, such as tool invocation, role-playing, and reasoning~\cite{wang_survey_2024, smart_yue, agentsurvey_Wang2024, cellagent_Xiao2024}. There is significant potential to bridge the gap between existing LLMs and their real-world applications by leveraging agent roles and scenario simulations\cite{agentsense_mou, electionsim_zhang, agenthospital_li}. Research such as AI Hospital~\cite{fan_ai_2024}, AgentClinic~\cite{schmidgall_agentclinic_2024}, and AIE~\cite{liao_automatic_2024} have created multi-agent simulation environments to evaluate the performance of LLMs in dynamic and interactive healthcare settings. Patient-$\Psi$~\cite{wang_patient-psi_2024} and CureFun~\cite{patientsimedu_li} propose approaches to train healthcare workers through simulated patient. However, these studies often neglect to simulate patient personality traits and demographic characteristics. Moreover, there is a lack of research utilizing scenario simulations to generate authentic data for improved real-world healthcare applications. Our work addresses these existing gaps and deficiencies, contributing to the advancement of more effective LLM-driven solutions in healthcare.

\section{Conclusion}

In this paper, we propose the following approach to contribute to a more efficient and patient-centered outpatient reception experience: (1) a system named Personalized Intelligent Outpatient Reception System (PIORS), to deliver personalized, high-quality, and efficient reception services. (2) a data generation framework named Service Flow aware Medical Scenario Simulation (SFMSS), to adapt LLMs to the real-world healthcare environments and support to the development of PIORS. Both automatic and human evaluation are conducted to demonstrate the effectiveness of PIORS and SFMSS in enhancing the application of LLMs in outpatient reception. Our framework has potential to transform the application of LLMs in outpatient reception, and be generalized to boarder real-world healthcare applications.

\section*{Limitations}
While PIORS and SFMSS demonstrates their effectiveness to contribute to a more efficient and patient-centered healthcare experience authentic, several limitations still remain. There may be regional variations in scenarios due to differences in language, culture, and social organization. As a result, relying solely on medical records from one hospital may limit the model's generalizability. Although the study includes diverse departments and personality simulations, the sample size of 2,400 is relatively small and may not fully capture the complexity of real clinical scenarios, particularly in departments focused on specific diseases. The simulated patient-nurse conversations may deviate from human-to-human interactions and require further validation. Additionally, PIORS have not been validated in real clinical environments. Despite these limitations, PIORS and SFMSS remains highly significant for the application of LLMs in healthcare settings, providing a novel method to enhance the application of LLMs in outpatient reception and boarder healthcare settings.

\section*{Ethics Statement}

\subsection*{Participant Recruitment}
We recruited participants of user study and expert evaluation through online advertising and networks of our co-authors, as well as snowball sampling. Users are adults with normal cognitive abilities, primarily consisting of university students and graduate students from various fields of study. Experts are individuals who have obtained a graduate (Master's or PhD) degree in clinical medicine or a related field, or who are currently pursuing such a degree, or who have at least two months of clinical work experience. We pay for each expert for participation. Users are volunteers. 

\subsection*{Informed Consent}

All participants in the user study and expert evaluation were 18 or older and provided informed consent. We did not assess any clinical outcomes. All data collected from the participants were de-identified and consented to be released for research purposes.

\subsection*{System and Data Usages} 

All the data and framework developed in this work are intended solely for academic research purposes. The framework developed in this work are intended to augment existing outpatient reception serve, not to replace it. Our framework is designed for academic and educational purposes only. Real-world deployment will require further work, including larger-scale training and testing, alignment with departmental and administrative information in real hospitals, and broader user and expert evaluations.

All the hospital outpatient records we utilize in the study are de-identified and consented to be released for research purposes, and do not contain and Personally Identifiable Information (PII) of any patients and hospital staff. The data has been anonymized and does not contain any sensitive information. It data is intended solely for academic research purposes.

\section*{Acknowledge}
The authors thank the doctors at the Eye \& ENT Hospital of Fudan University, the Shanghai Baoshan District Wusong Central Hospital and the Zhongshan Hospital for their valuable advice and support in system design and evaluation.

\bibliography{acl_latex}
\newpage
\appendix

\onecolumn
\section{Prompts Used in PIORS}
All the prompts provided are translated from the original Chinese version.
\subsection{PIORS-Nurse}
Here we provide the prompts used in PIORS-Nurse.
\paragraph{Query Generation Module} Below we provide the prompts used in the query generation module. First determine whether retrieval is necessary (upper), and then generate a query statement if required (lower).
\begin{figure}[htbp]
    \centering
    \includegraphics[width=0.9\linewidth]{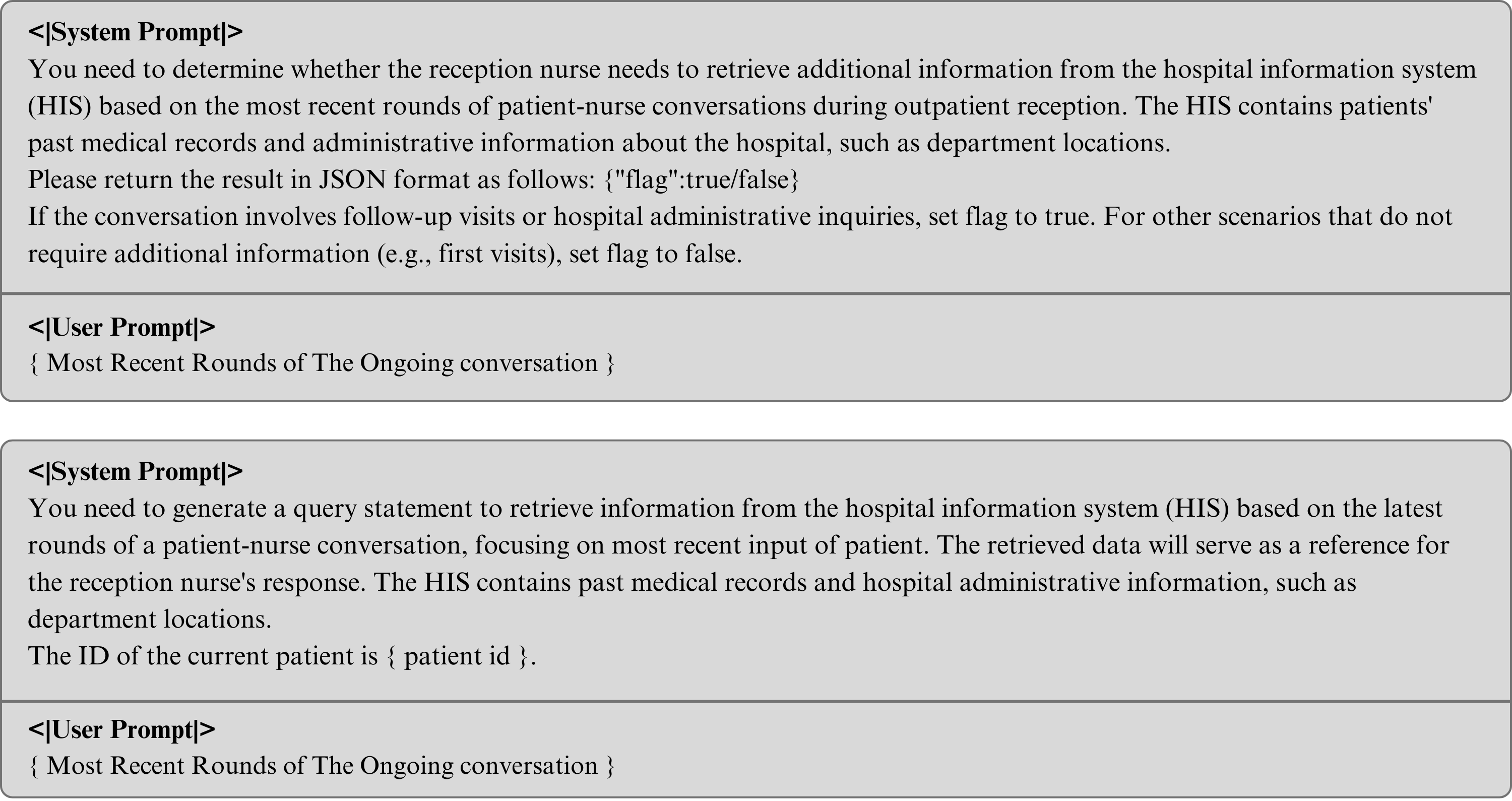}
\end{figure}
\paragraph{Summarization Module} Below we provide the prompts used in summarization module. Iteratively extract pre-diagnosis information (upper) and summarize for record creation (lower).

\begin{figure}[H]
    \centering
    \includegraphics[width=0.9\linewidth]{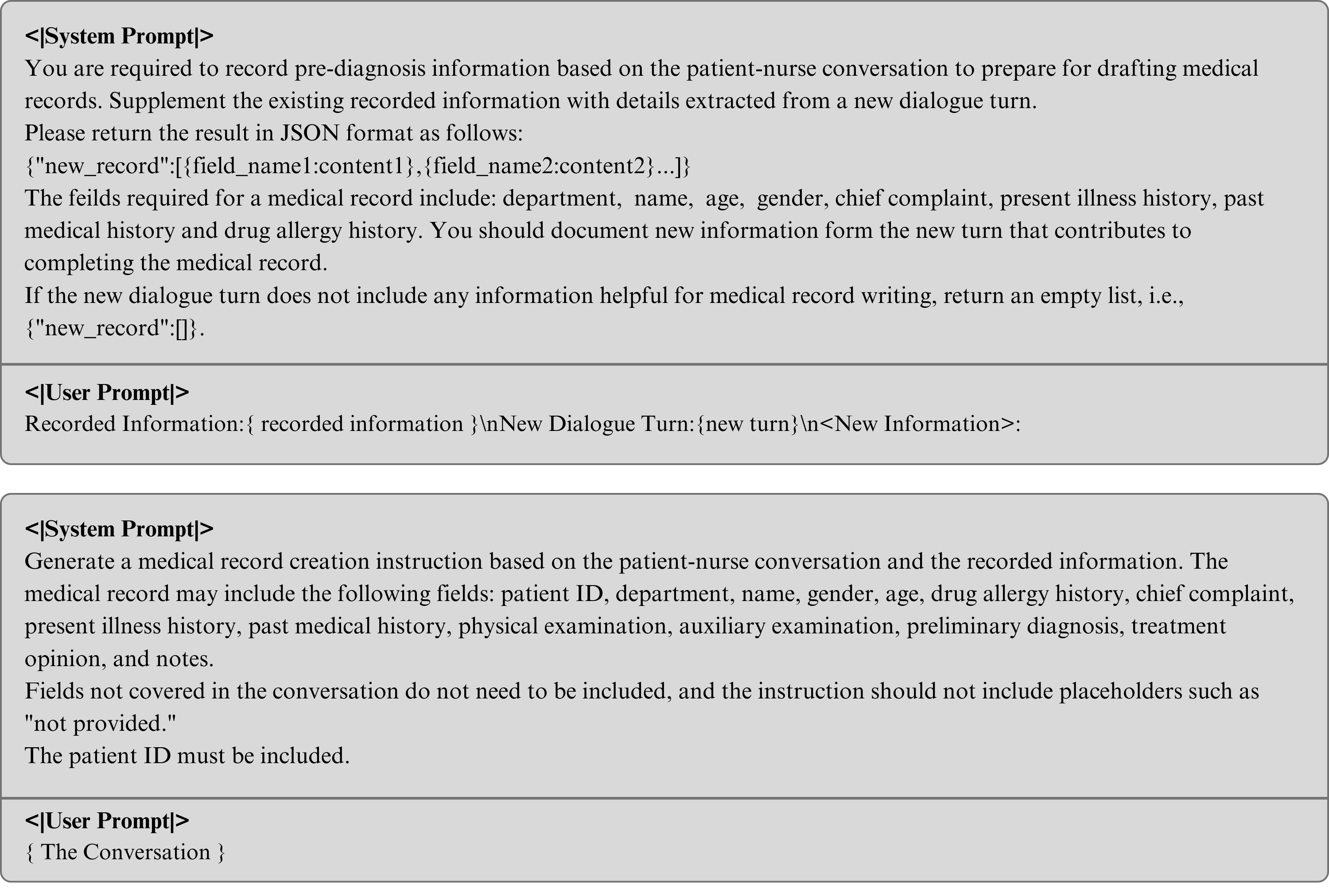}
\end{figure}

\newpage
\subsection{HospInfo-Assistant}
The prompts used in HospInfo-Assistant for extracting operation parameters are shown below.
\begin{figure}[htbp]
    \centering
    \includegraphics[width=1\linewidth]{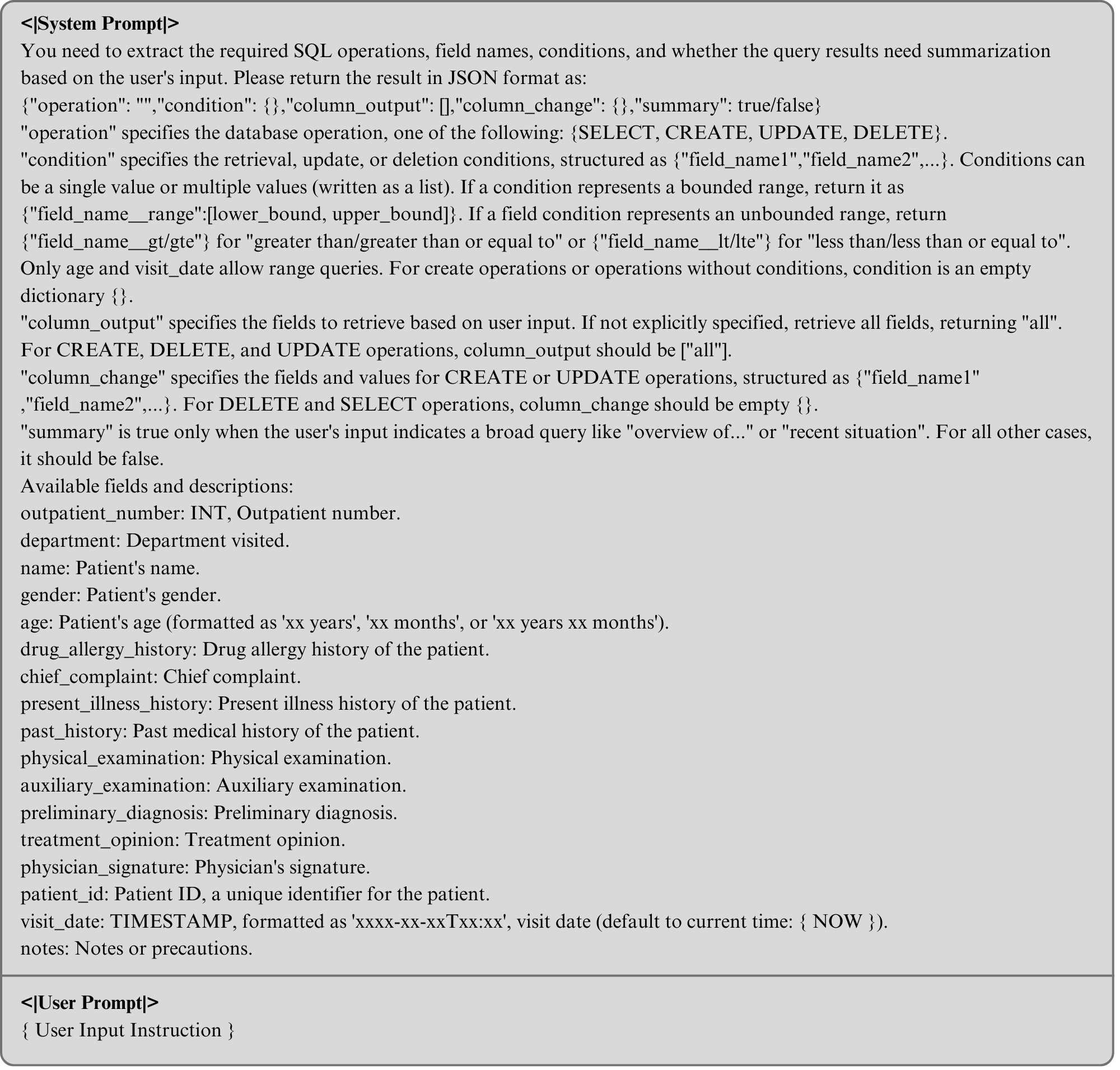}
\end{figure}
\newpage
\section{Prompts Used in SFMSS}
All the prompts provided are translated from the original Chinese version.
\label{appendix:SFMSS-prompts}
\subsection{Reception Nurse simulator}
Here we provide the prompts used in the reception nurse simulator. Below are the prompts for action decision (upper) and response generation (Lower). 
\begin{figure}[H]
    \centering
    \includegraphics[width=0.9\linewidth]{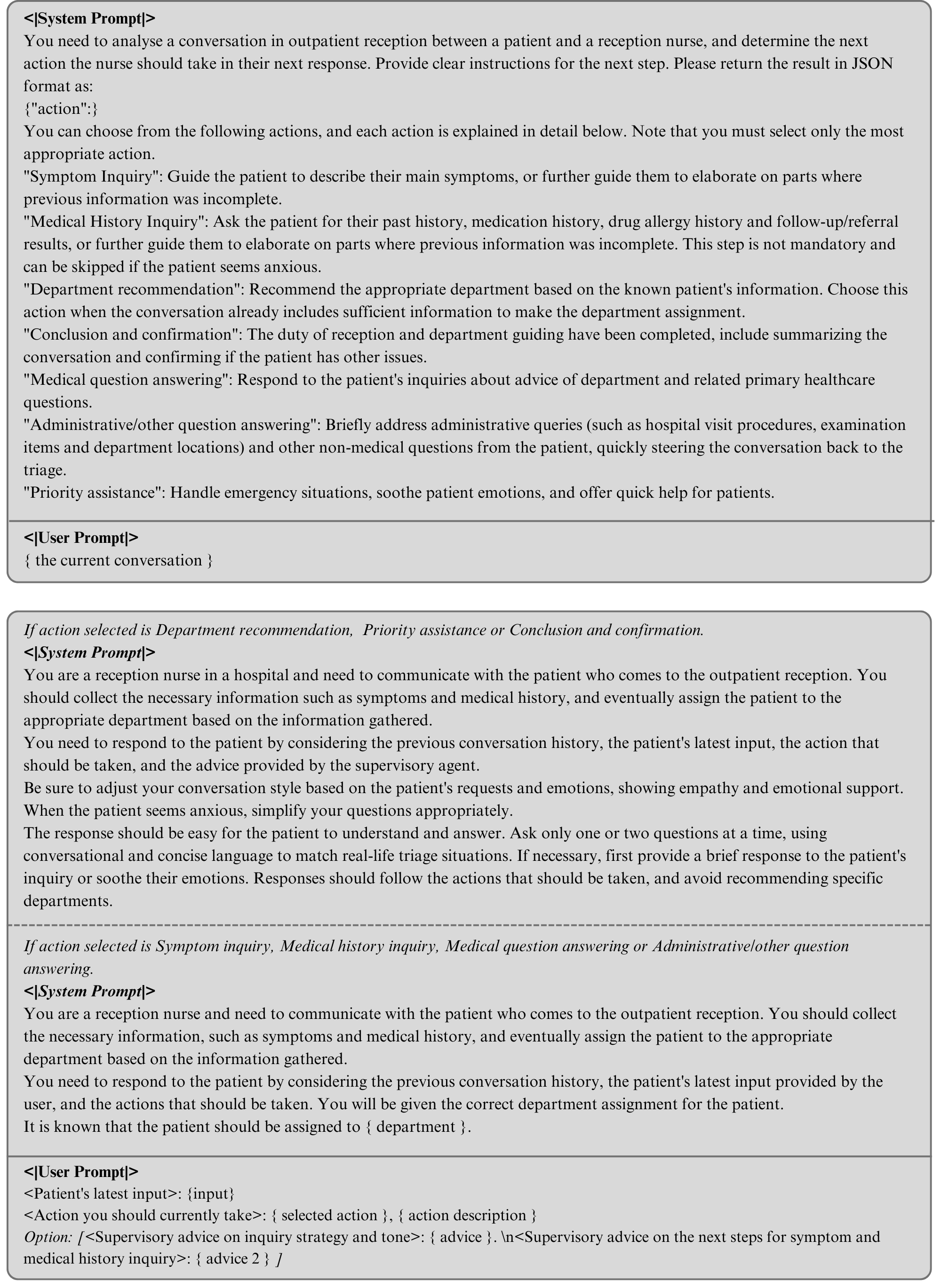}
    \label{fig:prompt_nurse}
\end{figure}
\newpage

\subsection{Patient simulator}

\textbf{Patient Profile Generating}

Prompts used for patient profile generating, including behavioural preference (Upper) and scene description (Lower).

\begin{figure}[H]
    \centering
    \includegraphics[width=0.9\linewidth]{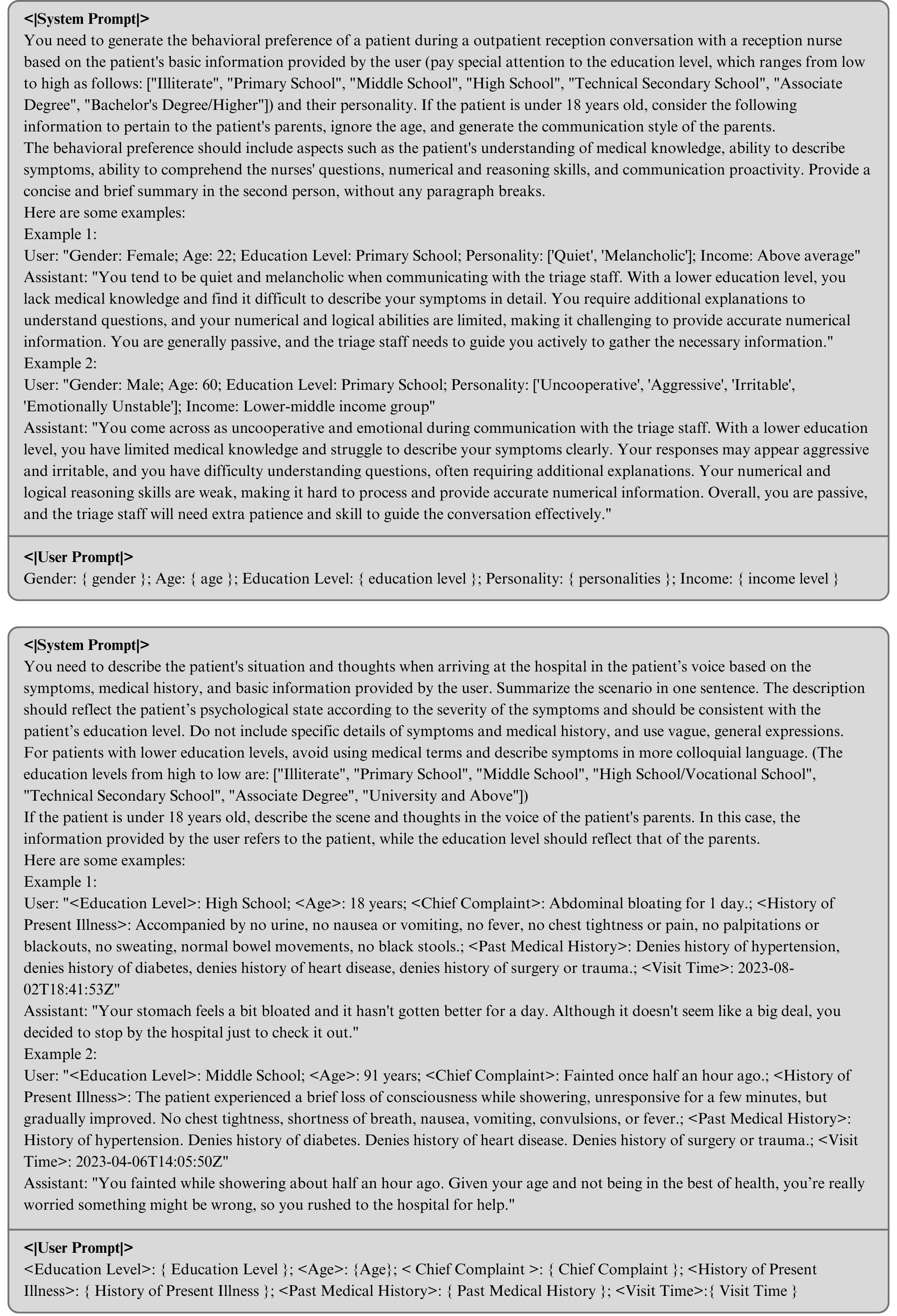}
    \label{fig:prompt_patient}
\end{figure}

\newpage

\textbf{Response Pipeline}

Prompts used inside the patient simulator, including action decision (upper) and response generation (Lower). 
\begin{figure}[H]
    \centering
    \includegraphics[width=0.9\linewidth]{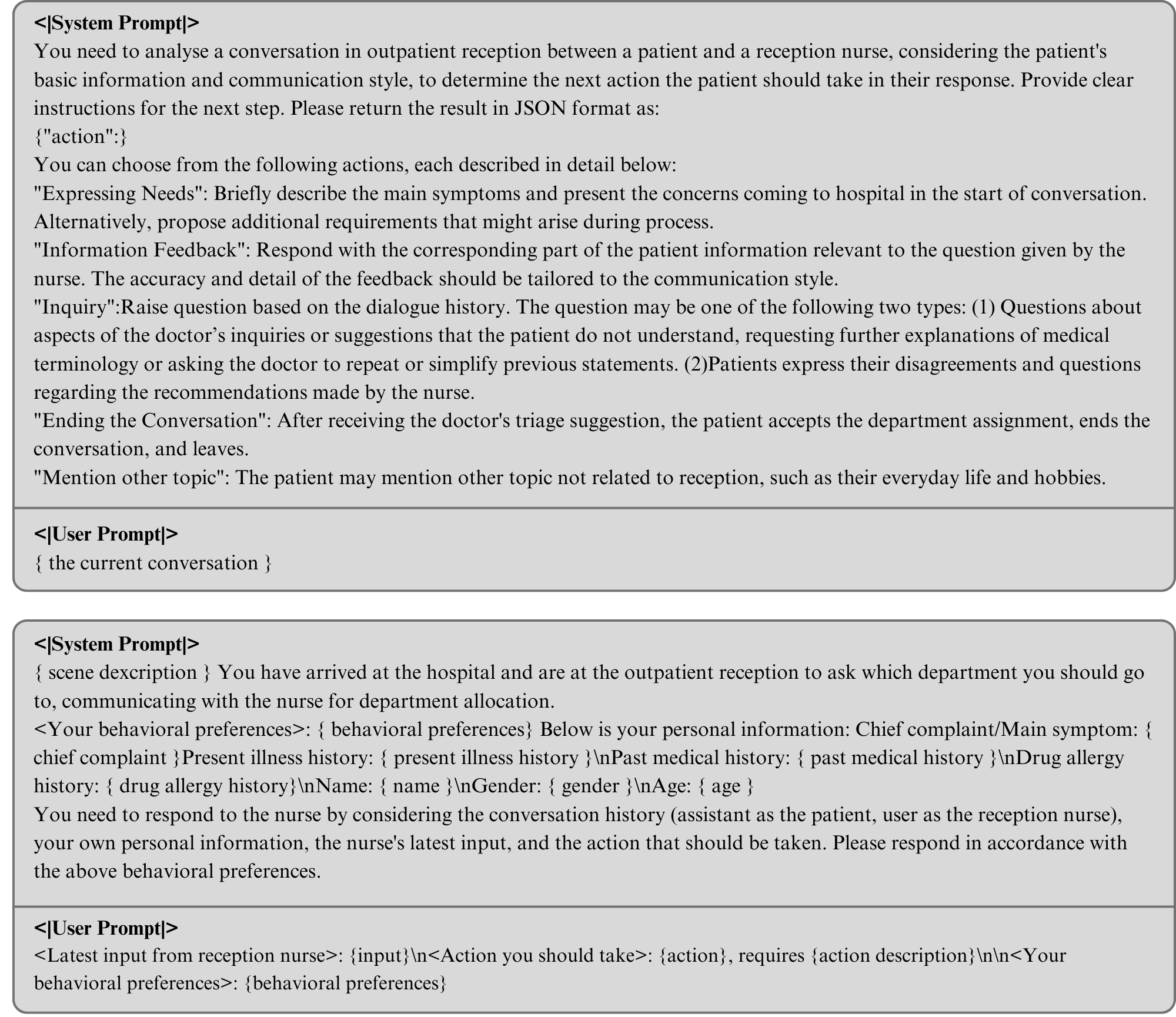}
\end{figure}

\subsection{Supervisor Agent}
\textbf{Dialogue quality}

Below is prompts used in dialogue quality supervisor.
\begin{figure}[H]
    \centering
    \includegraphics[width=0.9\linewidth]{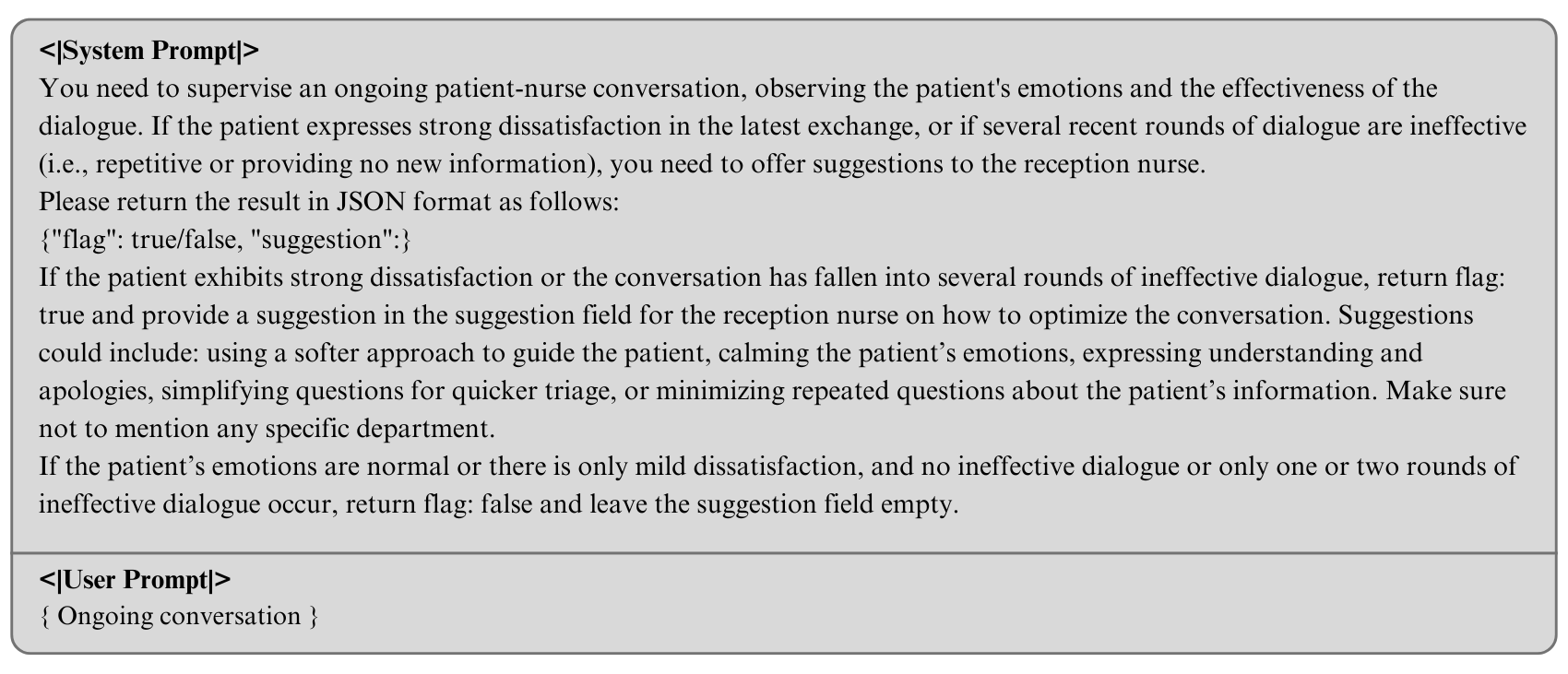}
\end{figure}

\newpage

\textbf{Information gathering}

Below is prompts used in dialogue quality supervisor.
\begin{figure}[H]
    \centering
    \includegraphics[width=0.9\linewidth]{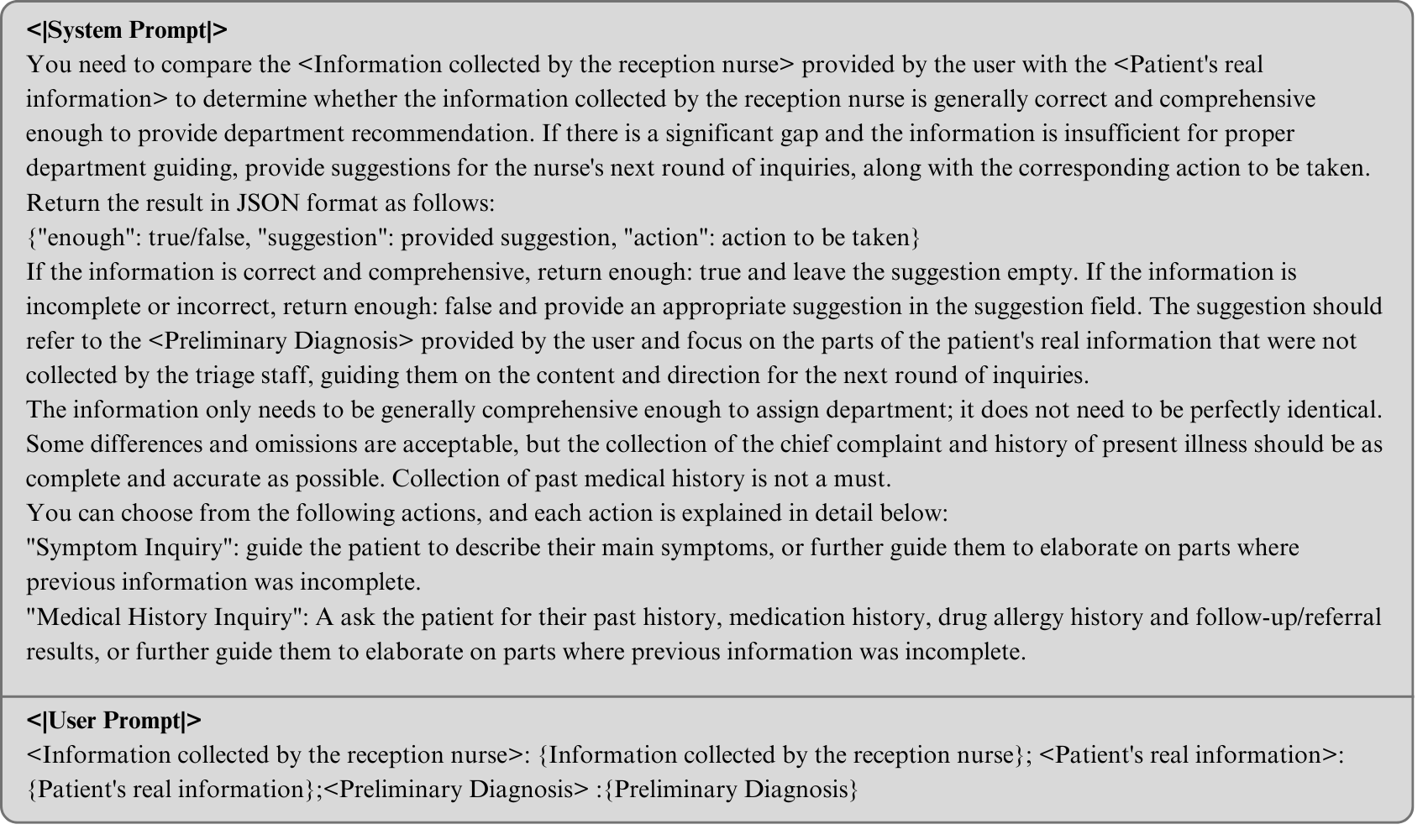}
\end{figure}

\section{Action Space Definition}
\label{appendix:action_space}
\subsection{First Visit Scenario}
In first visit scenarios, the action descriptions of reception nurse simulator and patient simulator is defined in Table\ref{tab:actino-doc} and Table \ref{tab:action-patient}.
\begin{table}[htbp]
\centering
\begin{tabular}{@{}cl@{}}
\toprule
\textbf{Nurse Actions}            & \multicolumn{1}{c}{\textbf{Actions Description}}                                                                                                                                                                                                                        \\ \midrule
Symptom Inquiry                   & Guide the patient to describe their main symptoms.                                                                                                                                                                                                                      \\
Medical History Inquiry           & \begin{tabular}[c]{@{}l@{}}Ask the patient for their past medical history, medication history, \\ drug allergy history and follow-up/referral results.\end{tabular}                                                                                                     \\
Department Recommendation         & \begin{tabular}[c]{@{}l@{}}Recommend the appropriate department based on the known \\ patient's information.\end{tabular}                                                                                                                                               \\
Priority Assistance               & \begin{tabular}[c]{@{}l@{}}Handle emergency situations, soothe patient emotions, and offer \\ quick help for patients.\end{tabular}                                                                                                                                     \\
Medical Question Answering        & \begin{tabular}[c]{@{}l@{}}Respond to the patient's inquiries about advice of department and \\ related primary healthcare questions.\end{tabular}                                                                                                                      \\
Administrative Question Answering & \begin{tabular}[c]{@{}l@{}}Address administrative queries (such as hospital visit procedures,\\  examination items and department locations) and other \\ non-medical questions from the patient, quickly steering the \\ conversation back to the triage.\end{tabular} \\
Conclusion and Confirmation       & \begin{tabular}[c]{@{}l@{}}Summarize the conversation and confirm if the patient has other \\ issues.\end{tabular}                                                                                                                                                      \\ \bottomrule
\end{tabular}
\caption{Definitions of reception nurse's actions in first visits.}
\label{tab:actino-doc}
\end{table}

\begin{table}[H]
\centering
\begin{tabular}{@{}cl@{}}
\toprule
\textbf{Patient Actions} & \multicolumn{1}{c}{\textbf{Actions Definition}}                                                                                                                                                                                                                                                                         \\ \midrule
Expressing Needs         & Describe main symptoms and concerns in the start of conversation.                                                                                                                                                                                                                                                       \\
Information Feedback     & \begin{tabular}[c]{@{}l@{}}Respond to the question raised by the nurse. The accuracy and detail of the\\ feedback should be tailored to the communication style. Responses may \\ include misunderstandings, answering questions not asked, partially \\ answering the questions, or not answering at all.\end{tabular} \\
Mention Other Topic      & \begin{tabular}[c]{@{}l@{}}Mention other topic not related to reception, such as their everyday life and \\ hobbies.\end{tabular}                                                                                                                                                                                       \\
Inquiry                  & \begin{tabular}[c]{@{}l@{}}Raise questions based on the dialogue history. Questions may be related \\ to nurse’s inquiries or suggestions that the patient do not understand or \\ disagree.\end{tabular}                                                                                                               \\
Ending the Conversation  & Confirm the recommended department and end the conversation.                                                                                                                                                                                                                                                            \\ \bottomrule
\end{tabular}
\caption{Definitions of patient's actions in first visits.}
\label{tab:action-patient}
\end{table}

\subsection{Follow-up Scenario}
In follow-up visit scenarios, we refine the action description of reception nurse and patient in Table\ref{tab:action-doc-f} and Table\ref{tab:action-patient-f}.

\begin{table}[htbp]
\centering
\begin{tabular}{@{}cl@{}}
\toprule
\textbf{Nurse Actions}            & \multicolumn{1}{c}{\textbf{Actions Description}}                                                                                                                                                                                                                        \\ \midrule
Symptom Inquiry                   & \begin{tabular}[c]{@{}l@{}}Guide the patient to describe changes in symptoms since their first \\ visit.\end{tabular}                                                                                                                                                   \\
Medical History Inquiry           & \begin{tabular}[c]{@{}l@{}}Ask the patient about any updates or additions to their past \\ medical history, adherence to treatment opinions form clinical \\ doctors, and examination results.\end{tabular}                                                             \\
Department Recommendation         & \begin{tabular}[c]{@{}l@{}}Recommend the appropriate department (usually the same) based \\ on the medical record of initial consultation.\end{tabular}                                                                                                                 \\
Priority Assistance               & \begin{tabular}[c]{@{}l@{}}Handle emergency situations, soothe patient emotions, and offer \\ quick help for patients.\end{tabular}                                                                                                                                     \\
Medical Question Answering        & \begin{tabular}[c]{@{}l@{}}Respond to the patient's inquiries about advice of department and \\ related primary healthcare questions.\end{tabular}                                                                                                                      \\
Administrative Question Answering & \begin{tabular}[c]{@{}l@{}}Address administrative queries (such as hospital visit procedures,\\  examination items and department locations) and other \\ non-medical questions from the patient, quickly steering the \\ conversation back to the triage.\end{tabular} \\
Conclusion and Confirmation       & \begin{tabular}[c]{@{}l@{}}Summarize the conversation and confirm if the patient has other \\ issues.\end{tabular}                                                                                                                                                      \\ \bottomrule
\end{tabular}
\caption{Descriptions of reception nurse's actions in follow-up visits.}
\label{tab:action-doc-f}
\end{table}

\begin{table}[H]
\centering
\begin{tabular}{@{}cl@{}}
\toprule
\textbf{Patient Actions} & \multicolumn{1}{c}{\textbf{Actions Definition}}                                                                                                                                                                                                                                                                         \\ \midrule
Expressing Needs         & Propose follow-up needs based on the symptoms from the initial record.                                                                                                                                                                                                                                                  \\
Information Feedback     & \begin{tabular}[c]{@{}l@{}}Respond to the question raised by the nurse. The accuracy and detail of the\\ feedback should be tailored to the communication style. Responses may \\ include misunderstandings, answering questions not asked, partially \\ answering the questions, or not answering at all.\end{tabular} \\
Mention Other Topic      & \begin{tabular}[c]{@{}l@{}}Mention other topic not related to reception, such as their everyday life and \\ hobbies.\end{tabular}                                                                                                                                                                                       \\
Inquiry                  & \begin{tabular}[c]{@{}l@{}}Raise questions based on the dialogue history. Questions may be related \\ to nurse’s inquiries or suggestions that the patient do not understand or \\ disagree.\end{tabular}                                                                                                               \\
Ending the Conversation  & \begin{tabular}[c]{@{}l@{}}The department recommendation has already given. The nurse's inquiries \\ have concluded, or the patient no longer wishes to continue the conversation.\end{tabular}                                                                                                                         \\ \bottomrule
\end{tabular}
\caption{Descriptions of Patient's actions in follow-up visits.}
\label{tab:action-patient-f}
\end{table}

\section{Personality simulation}
\label{appendix:personality}
\begin{table}[htbp]
\centering
\begin{tabular}{@{}lll@{}}
\toprule
Big Five Trait                  & High Marker                                                                                                                 & Low Marker                                                                                                                              \\ \midrule
\textbf{Extraversion}           & \begin{tabular}[c]{@{}l@{}}Outgoing, Talkative, \\ Bold/Confident, Positive, \\ Energetic, Optimistic/Cheerful\end{tabular} & \begin{tabular}[c]{@{}l@{}}Introverted, Silent, Timid/Unconfident, \\ Negative, Lacking Energy, Melancholy\end{tabular}                 \\
\textbf{Agreeableness}          & \begin{tabular}[c]{@{}l@{}}Friendly, Trusting, Cooperative, \\ Humble, Easygoing\end{tabular}                               & \begin{tabular}[c]{@{}l@{}}Aggressive, Distrustful, Dishonest, \\ Uncooperative, Arrogant, \\ Unaccommodating\end{tabular}              \\
\textbf{Conscientiousness}      & Organized, Diligent, Thorough                                                                                               & Disorganized, Careless, Forgetful                                                                                                       \\
\textbf{Openness to Experience} & \begin{tabular}[c]{@{}l@{}}Imaginative, Creative, Reflective, \\ Emotionally Sensitive, \\ Curious, Analytical\end{tabular} & \begin{tabular}[c]{@{}l@{}}Unimaginative, Uncreative, Unreflective, \\ Emotionally Closed, Uncurious\end{tabular}                       \\
\textbf{Neuroticism}            & Calm, Patient, Emotionally Stable                                                                                           & \begin{tabular}[c]{@{}l@{}}Tense, Anxious, Worrisome, Irritable,\\ Impulsive, Easily Dissatisfied, \\ Emotionally Unstable\end{tabular} \\ \bottomrule
\end{tabular}
\label{tab:bigfive_adjs}
\caption{High and low markers pre-defined for BigFive personality traits.}
\end{table}

\twocolumn
\section{Experiment Setup Details}
\subsection{Outpatient Records Details}
\label{appendix:record_details}
\begin{figure}[htbp]
    \centering
    \includegraphics[width=1\linewidth]{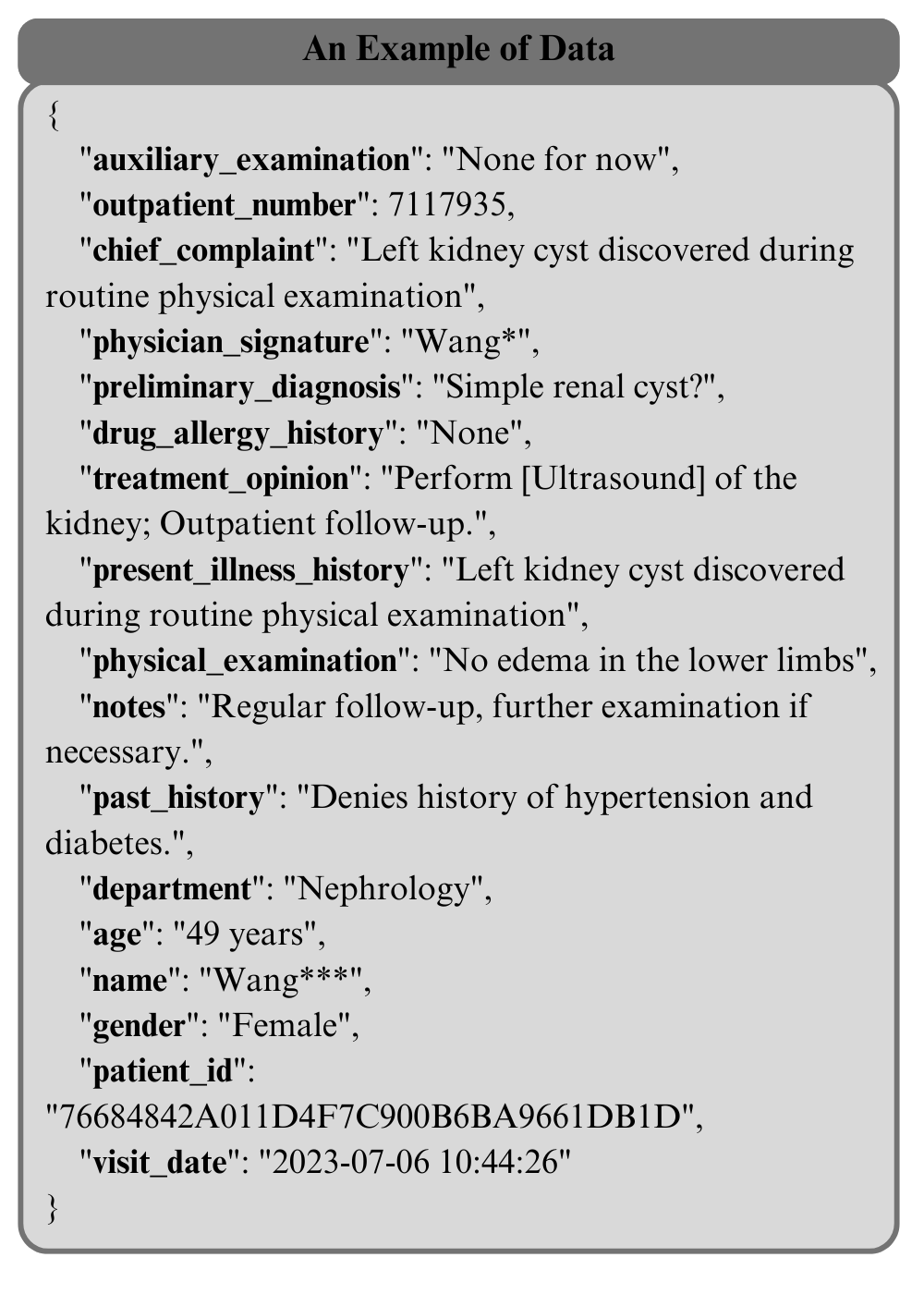}
    \caption{An example outpatient medical record.}
    \label{fig:record_example}
\end{figure}

Each record consists of 17 fields, the field names and their definitions are as follows:

\textbf{\textit{Outpatient Number}}: The unique id of an outpatient medical record.\textbf{ \textit{Chief Complaint}}: The primary reason or main symptom for which a patient seeks medical care. \textbf{\textit{History of Present Illness}}: A detailed account of the patient's current symptoms and medical condition. \textbf{\textit{Past Medical History}}: A comprehensive summary of a patient's previous health conditions, medical treatments, surgeries, hospitalizations, chronic illnesses, allergies, and medications. It may also include significant family medical history and lifestyle factors like smoking or alcohol use. \textbf{\textit{Department}}: The specific unit or division within the hospital where a patient is treated. \textbf{\textit{Drug Allergy History}}: A patient's previous allergic reactions to medications. \textbf{\textit{Age}}: The age of the patient. \textbf{\textit{Gender}}: The gender of the patient. \textbf{\textit{Name}}: The anonymized name of the patient. \textbf{\textit{Visit Time}}: The registration time of the patient's current visit. \textit{Patient ID}: The unique ID to identify the patient. \textbf{\textit{Preliminary Diagnosis}}: The initial assessment or hypothesis made by doctors. \textbf{\textit{Physical Examination}}: The systematic evaluation of a patient's body by the doctor. \textbf{\textit{Auxiliary examination}}: Additional diagnostic tests or procedures that are used to support or confirm a diagnosis. \textbf{\textit{Notes}}: Specially notation given by doctors. \textbf{\textit{Physician signature}}: The official signature provided by the attending doctor who is responsible for the patient's care. This field has been anonymized. \textbf{\textit{Treatment opinion}}: The professional recommendations or suggestions provided by a doctor for the patient.

An example record in the dataset is shown in Figure \ref{fig:record_example}.

\subsection{Prompts for Baselines}
\label{appendix:prompt_baseline}
\begin{figure}[htbp]
    \centering
    \includegraphics[width=1\linewidth]{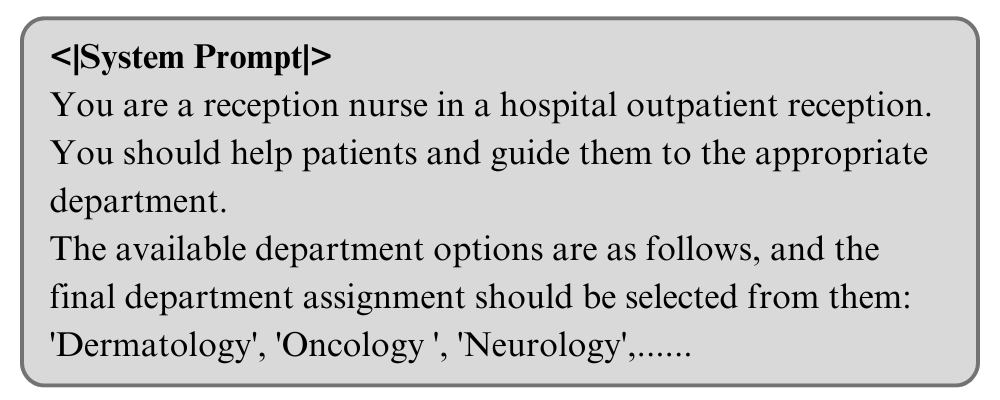}
    \caption{Role-play prompt for directly prompted baselines and evaluation.}
    \label{fig:prompt_eval}
\end{figure}

In the automatic evaluation, models are prompted to role-play the reception nurse, with the system prompt in Figure\ref{fig:prompt_eval}. The \textbf{directly prompted baselines} also share the same instruction.

\begin{figure}[htbp]
    \centering
    \includegraphics[width=1\linewidth]{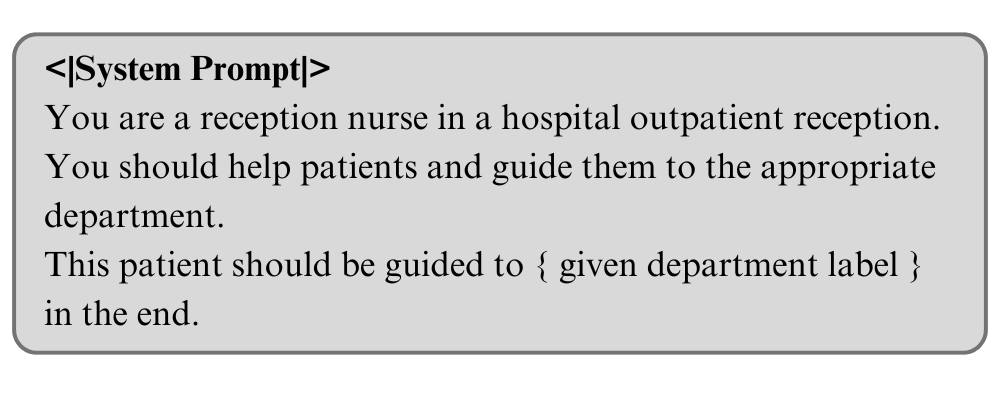}
    \caption{Prompt for normal role-playing data generation method.}
    \label{fig:prompt_baseline}
\end{figure}

When training the \textbf{SF-ablated nurse}, we directly prompt GPT-4o as reception nurse. The prompt is provided in Figure \ref{fig:prompt_baseline}.

All the prompts provided are translated from the original Chinese version.
\section{Evaluation Details}
\subsection{Prompts for Automatic Evaluation}
\begin{figure}[htbp]
    \centering
    \includegraphics[width=1\linewidth]{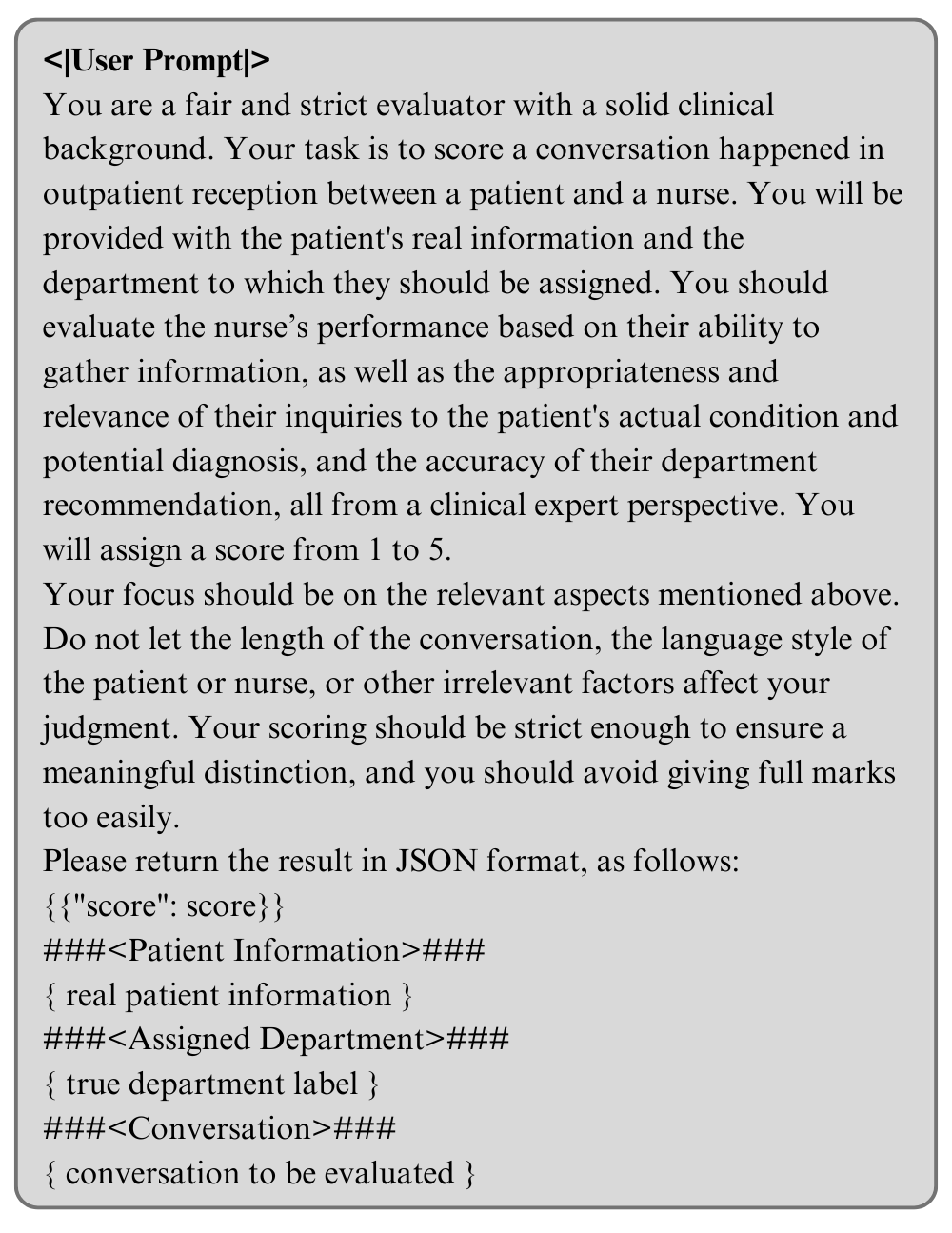}
    \caption{Instruction for GPT-4o to provide the Overall Score.}
    \label{fig:overall_eval_prompt}
\end{figure}

\begin{figure}[htbp]
    \centering
    \includegraphics[width=1\linewidth]{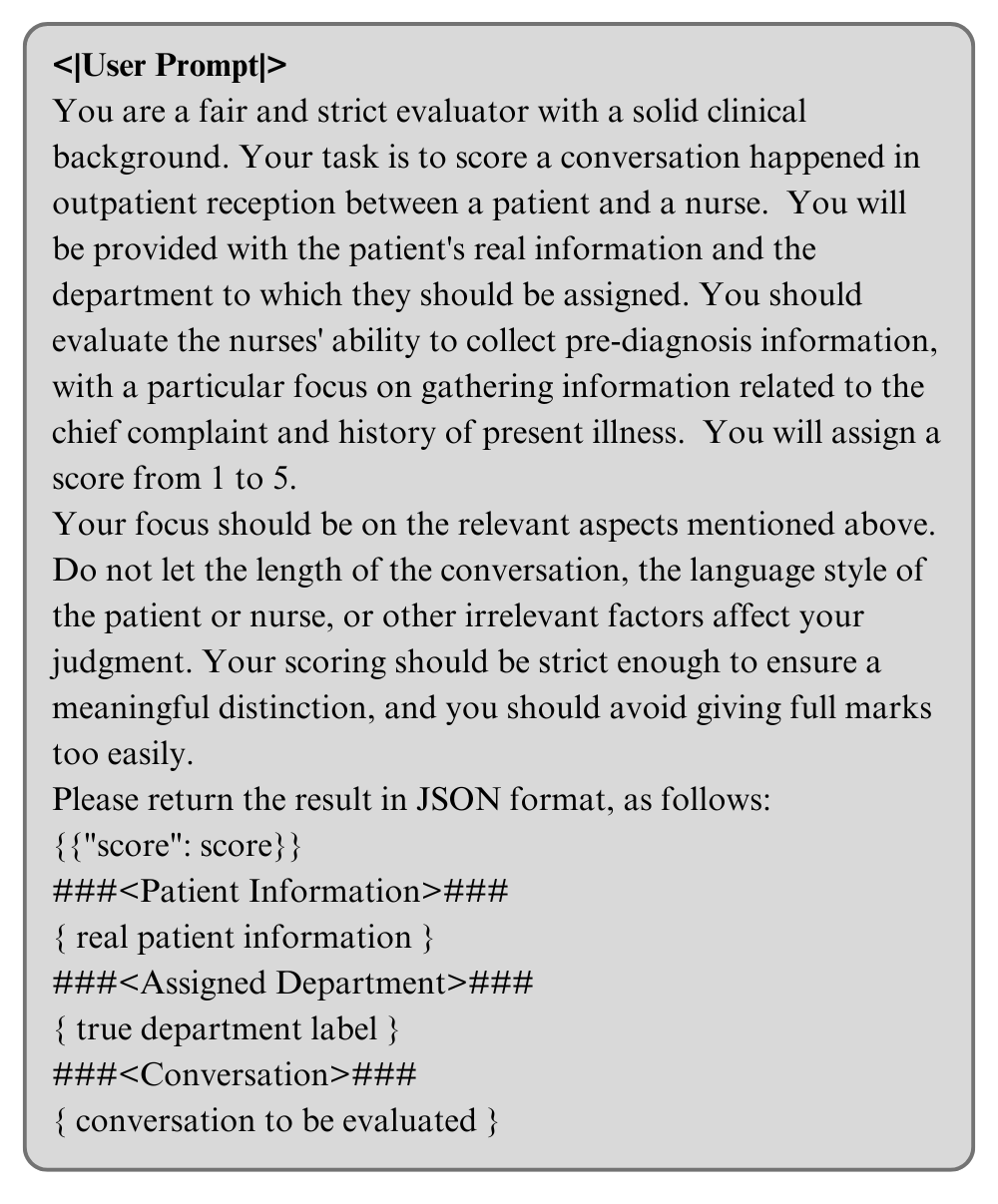}
    \caption{Instruction for GPT-4o to provide the Info. Score.}
    \label{fig:info_eval_prompt}
\end{figure}

Here we provide our prompts for automatic evaluation in Figure \ref{fig:overall_eval_prompt} and Figure \ref{fig:info_eval_prompt}. (All the prompts provided are translated from the original Chinese version.)

\subsection{Expert Evaluation Details}
\label{fidelity_levels}
\textbf{Fidelity Options}

\textit{Extremely High}: Fully consistent with the behavior of real clinical patients. The speech and behavior logic are indistinguishable from real patients, aligning with the characteristics of real patient groups.

\textit{High}: Closely resemble the behavior and performance of real clinical patients, with only slight deviations. Overall, the simulation maintains the logical consistency and characteristics of real patient groups.

\textit{Moderate}: The simulated patient's behavior differs noticeably from real clinical patients. The speech and behavior logic appear unnatural, making it difficult to reflect the characteristics of real patient groups.

\textit{Low}: The simulated patient's speech and behavior are significantly different from those of real clinical patients. The performance shows clear logical inconsistencies and behavioral issues, resulting in a lack of realism.

\section{Additional Automatic Evaluation Results}
\label{appendix:add_results}
\begin{figure}[htbp]
    \centering
    \includegraphics[width=0.9\linewidth]{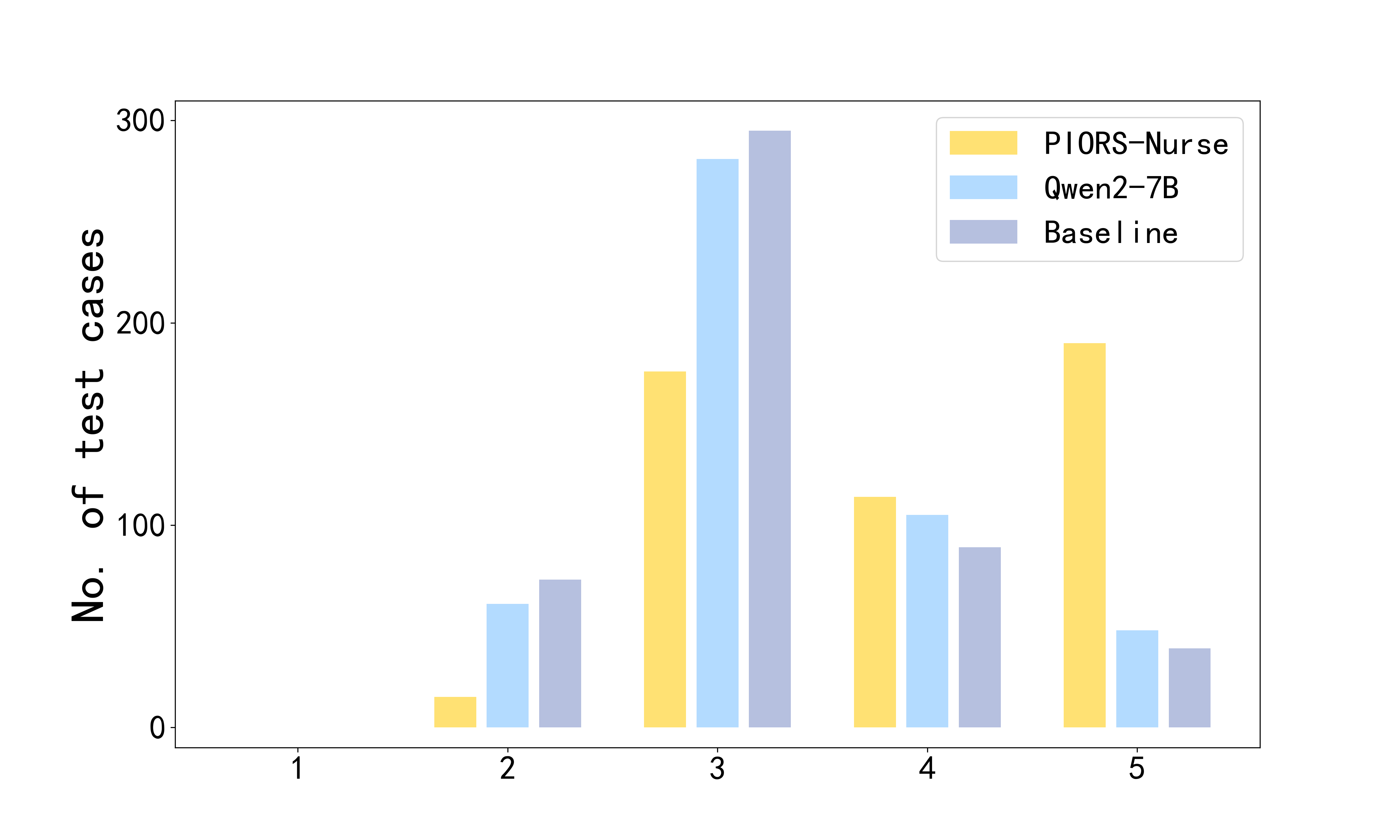}
    \caption{Info Score distribution of different models.}
    \label{fig:info}
\end{figure}

\begin{figure}[htbp]
    \centering
    \includegraphics[width=0.9\linewidth]{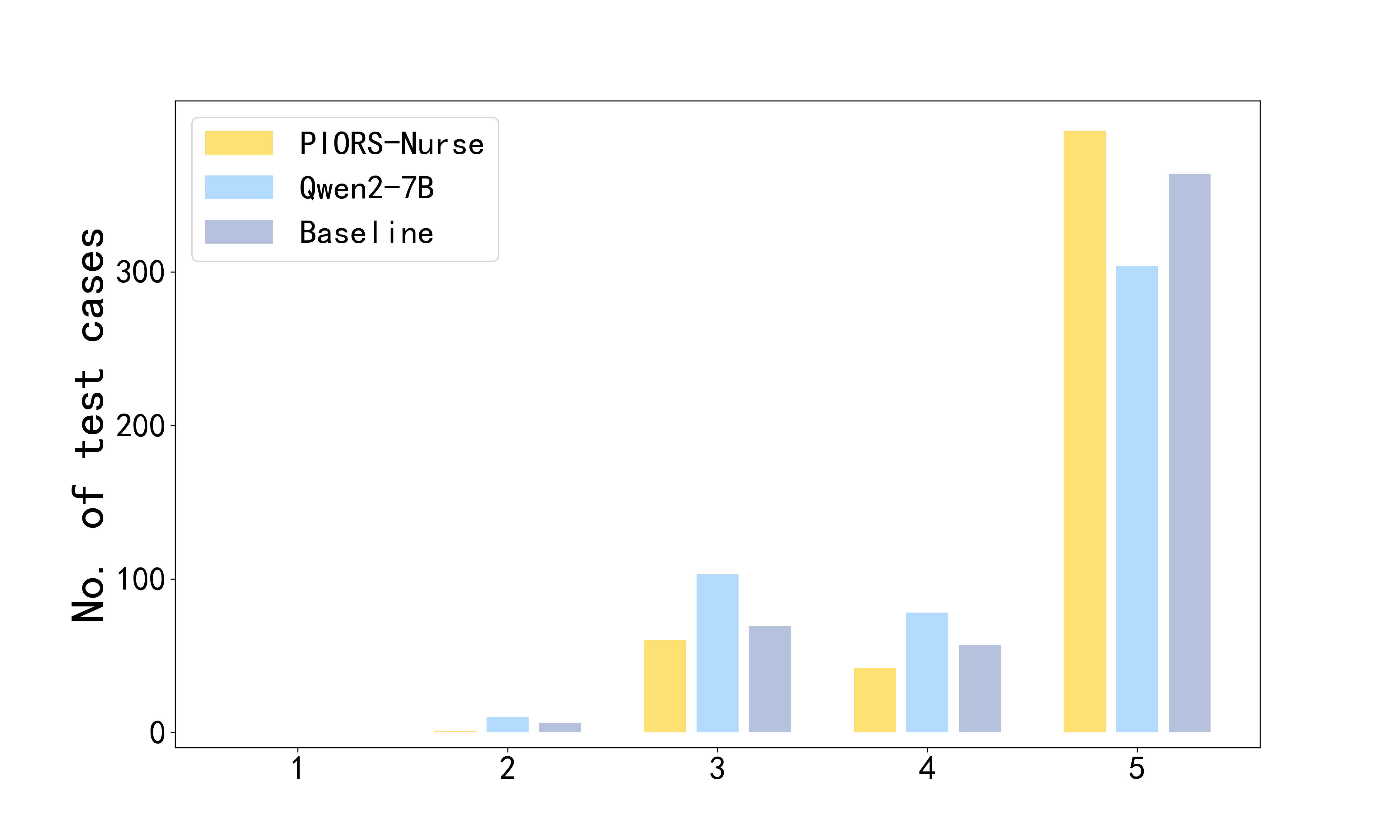}
    \caption{Overall Score distribution of different models.}
    \label{fig:overall_score}
\end{figure}


\begin{figure}[htbp]
    \centering
    \includegraphics[width=1\linewidth]{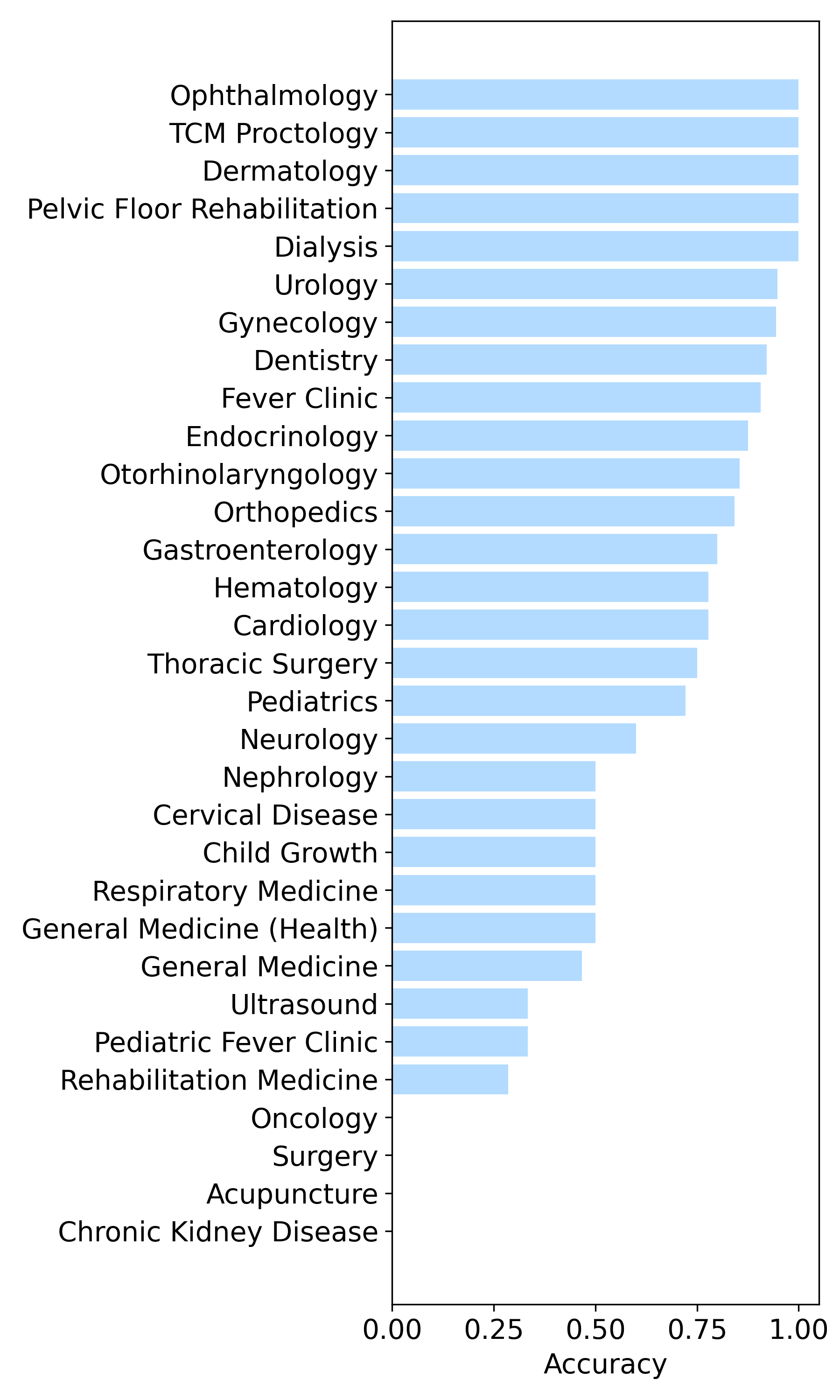}
    \caption{PASS-Nurse test results: Accuracy grouped by different departments.}
    \label{fig:department_acc}
\end{figure}


\end{document}